\documentclass[journal]{IEEEtran}

\usepackage[pdftex]{graphicx}
\usepackage{amsmath}
\usepackage{amssymb}
\usepackage{amsfonts}

\usepackage{multirow} 
\usepackage{rotating}

\usepackage{subfig}
\usepackage{color}
\usepackage{ogonek}
\usepackage{float}
\usepackage{url}

\usepackage{tikz}
\usetikzlibrary{positioning}
\usetikzlibrary{arrows}
\tikzset{
    >=stealth',
    blackarrow/.style={
           ->,
           thick,
           shorten <=2pt,
           shorten >=2pt,}
}

\begin{document}




\title{Dissimilarity-based Ensembles for\\ Multiple Instance Learning}


\author{Veronika~Cheplygina,
        David~M.J.~Tax,
        and~Marco~Loog
\thanks{The authors are with the Pattern Recognition Laboratory, Delft University of Technology, The Netherlands. E-mail: v.cheplygina@tudelft.nl}
}

\maketitle

\begin{abstract}

In multiple instance learning, objects are sets (bags) of feature vectors (instances) rather than individual feature vectors. In this paper we address the problem of how these bags can best be represented. Two standard approaches are to use (dis)similarities between bags and prototype bags, or between bags and prototype instances. The first approach results in a relatively low-dimensional representation determined by the number of training bags, while the second approach results in a relatively high-dimensional representation, determined by the total number of instances in the training set. In this paper a third, intermediate approach is proposed, which links the two approaches and combines their strengths. Our classifier is inspired by a random subspace ensemble, and considers subspaces of the dissimilarity space, defined by subsets of instances, as prototypes. We provide guidelines for using such an ensemble, and show state-of-the-art performances on a range of multiple instance learning problems. 

\end{abstract}


%
\IEEEpeerreviewmaketitle

\section{Introduction}\label{sec:intro}

\IEEEPARstart{N}{o}wadays, many applications face the problem of using weakly labeled data for training a classifier. For example, in image classification, we may only have an overall label for an image (such as a ``tiger''), not where the tiger is actually located in the image. Such problems are often formulated as multiple instance learning (MIL)~\cite{dietterich1997solving} problems. MIL is an extension of supervised learning, and occurs in cases when class labels are associated with sets (\emph{bags}) of feature vectors (\emph{instances}) rather than with individual feature vectors. The bag labels provide weak information about the instance labels. For example, the label ``tiger'' could apply to only some of the image patches, because patches of sand, sky or other surroundings could be present as well. This is a natural representation for many real-world problems, therefore MIL has been successfully used to address molecule~\cite{dietterich1997solving} or drug~\cite{fu2012implementation} activity prediction, image classification~\cite{andrews2002multiple,chen2006miles}, document categorization~\cite{zhou2009multi}, computer aided diagnosis~\cite{fung2007multiple} and many other problems. 


We can group methods that learn in this weakly supervised setting in two categories. The first category, which we call instance-based methods, relies on assumptions~\cite{foulds2010review} about the labels to recover the instance labels, and classifies a bag by first classifying that bag's instances, and then fusing these outputs into a bag label. For example, the standard assumption is that a bag is positive if and only if at least one of its instances is positive. The second category, which we call bag-based methods, often use the collective assumption: they assume all instances contribute to the bag label, and that bags with the same label are somehow similar to each other. Therefore, bags can be classified directly using distances~\cite{wang2000solving} or kernels~\cite{gartner2002multi}, or by converting bags to a single-instance representation and using supervised learners~\cite{chen2006miles,tax2011bag}. The bag-based methods have frequently demonstrated the best performances on a wide range of datasets.

One of the ways to represent structured objects, such as bags, in a feature space, is to describe them relative a set of reference objects or prototypes. This novel approach is called the dissimilarity representation~\cite{pekalska2005dissimilarity} and is in contrast to traditional pattern recognition because the dimensions of this space are defined in a relative way: the dissimilarity to the $j$-th prototype can therefore be seen as the $j$-th feature in the transformed space. A successful approach we studied, uses training bags as prototypes~\cite{tax2011bag,cheplygina2012class}, while an alternative approach in~\cite{chen2006miles} uses all the training instances as prototypes. Both alternatives have demonstrated the best performances on a wide range of MIL problems, and, as we show in this paper, are in fact strongly related. However, both approaches are extremes with respect to the dimensionality and the information content of the resulting dissimilarity space. The bag representation reduces the dimensionality from the number of instances to the number of bags, but risks losing information. The instance representation preserves more information, but increases the dimensionality dramatically, possibly including many redundant features. 

Consider each prototype as a different information source. The bag and instance representations are analogous to two different ways --- concatenating and averaging --- of combining such information sources~\cite{pekalska2001combining}. In general pattern recognition problems there is a third alternative: ensemble methods~\cite{kittler1998combining,duin2000experiments}. When presented with different information sources, we can train a classifier on each information source, and combine the classifier decisions in the test phase. 
For dissimilarities, this translates into training a classifier on a different subspace of the dissimilarity space, where each subspace corresponds to a a subset of bags or instances. Such an ensemble is in between the two opposing representations, and 
offers several advantages: the information provided by different dissimilarities is preserved, the dimensionality of each subspace is lower. Therefore the ensemble has the potential to be more robust than a single classifier. 

We introduced dissimilarity-based subspace ensembles in~\cite{cheplygina2013combining}, however, the preliminary results did not meet our expectations because the ensembles typically did not outperform the single classifier dissimilarity representations. We describe the ensembles in Section 3 and investigate these methods in more depth to provide understanding of the relationship between ensemble design and ensemble performance, and therefore of our earlier results. In Section 4 we show significantly improved, competitive results on many MIL datasets. Furthermore, our results provide insight the structure of some popular MIL problems, and into the success of the dissimilarity space for these problems.

\section{Dissimilarity-based Multiple Instance Learning}

\subsection{Data Representation}

In multiple instance learning, an object is represented by a bag $B_i  = \{\mathbf{x}_{ik}| k=1,...,n_i\} \subset \mathbb{R}^d$ of $n_i$ feature vectors or instances. The training set $\mathcal{T} = \{(B_i, y_i) | i=1,...N\}$ consists of positive ($y_i = +1$) and negative ($y_i = -1$) bags, although multi-class extensions are also possible~\cite{zhou2006multi}. The standard assumption for MIL is that there are instance labels $y_{ik}$ which relate to the bag labels as follows: a bag is positive if and only if it contains at least one positive, or \emph{concept} instance: $y_i = \max_k y_{ik}$. In this case, it might be worthwhile to search for only these informative instances. Alternative formulations, where a fraction or even all instances are considered informative, have also been proposed~\cite{foulds2010review}. 


We can represent an object, and therefore also a MIL bag $B_i$, by its dissimilarities to prototype objects in a representation set $\mathcal{R}$~\cite{pekalska2005dissimilarity}. Often $\mathcal{R}$ is taken to be a subset of size $M$ of the training set $\mathcal{T}$ of size $N$ ($M \leq N$). If we apply this to MIL, each bag is represented as $\mathbf{d}(B_i, \mathcal{T})= [d(B_i, B_1), ... d(B_i, B_M)]$: a vector of $M$ dissimilarities. Therefore, each bag is represented by a single feature vector $\mathbf{d}$ and the MIL problem can be viewed as a standard supervised learning problem.

The bag dissimilarity $d(B_i, B_j)$ is defined as a function of the pairwise instance dissimilarities $[d({\mathbf{x}}_{ik}, {\mathbf{x}}_{jl})]_{n_i \times n_j}$.  There are many alternative definitions (see ~\cite{tax2011bag,cheplygina2012does}) but in this work we focus on the average minimum instance distance, which tends to perform well in practice. Suppose that we are only given one prototype $B_j$. With the proposed bag dissimilarity, the bag representation of $B_i$ using prototype $B_j$ is:

\begin{equation}\label{eq:meanmin}
d^{bag}(B_i,B_j) = \frac{1}{n_i}\sum_{k=1}^{n_i} \min_{l} d(\mathbf{x}_{ik}, \mathbf{x}_{jl})
\end{equation}

Note that the dissimilarity between bag $B_i$ and $B_j$ is now reduced to a scalar, and $\mathbf{d}(B_i,\mathcal{T})$ becomes an $M$-dimensional vector.

A related method, MILES \cite{chen2006miles}, considers a different definition of prototypes, by using the training instances rather than the training bags. The motivation is that, when we assume just a few concept instances per bag, it is better to consider just these informative instances rather than the bag as a whole. MILES is originally a similarity-based approach, where the similarity is defined as  $s(B_i,\mathbf{x}) = \max_{k} \exp{(-\frac{d(\mathbf{x}_{ik}, \mathbf{x})}{\sigma^2})}$ and $\sigma$ is the parameter of the radial basis function kernel. However, by leaving out kernel and the need to choose $\sigma$, we get a dissimilarity-based counterpart. The instance representation of $B_i$ using the instances of $B_j$ is then defined as:

\begin{multline}\label{eq:rep_inst}
\mathbf{d}^{inst}(B_i,B_j) = [\min_{l} d(\mathbf{x}_{i1}, \mathbf{x}_{jl}), \min_{l} d(\mathbf{x}_{i2}, \mathbf{x}_{jl}), \cdots,\\ \min_{l} d(\mathbf{x}_{in_i}, \mathbf{x}_{jl})]
\end{multline}

Now the dissimilarity between $B_i$ and $B_j$ is summarized in a $n_i$-dimensional vector, resulting in a representation $\mathbf{d}(B_i,\mathcal{T})$ that has a dimensionality of $\sum_{k=1}^M n_k$. 

From this point onwards, we will discuss the dissimilarity matrices $D^{bag}$ and $D^{inst}$, which look as follows:

\vspace{0.7cm}

\scriptsize
$D^{bag}=$
\begin{equation}
 \begin{pmatrix}
  d^{bag}(B_1,B_1) & d^{bag}(B_1,B_2) & \cdots & d^{bag}(B_1,B_M) \\
  d^{bag}(B_2,B_1) & d^{bag}(B_2,B_2) & \cdots & d^{bag}(B_2,B_M) \\
  \vdots  & \vdots  & \ddots & \vdots  \\
  d^{bag}(B_N,B_1) & d^{bag}(B_N,B_2) & \cdots & d^{bag}(B_N,B_M)
 \end{pmatrix}
\end{equation}

\vspace{0.7cm}

%

$D^{inst}=$
\begin{equation}
 \begin{pmatrix}
 \mathbf{d}^{inst}(B_1,B_1) & \mathbf{d}^{inst}(B_1,B_2) & \cdots & \mathbf{d}^{inst}(B_1,B_M) \\
  \mathbf{d}^{inst}(B_2,B_1) & \mathbf{d}^{inst}(B_2,B_2) & \cdots & \mathbf{d}^{inst}(B_2,B_M) \\
  \vdots  & \vdots  & \ddots & \vdots  \\
  \mathbf{d}^{inst}(B_N,B_1) & \mathbf{d}^{inst}(B_N,B_2) & \cdots & \mathbf{d}^{inst}(B_N,B_M)  
  
   \end{pmatrix} 
   \end{equation}

\normalsize
\vspace{0.3cm}

$D^{bag}$ and $D^{inst}$ are two extremes with respect to the amount of information that is preserved. In cases where only a few instances per bag are informative, $D^{bag}$ could suffer from averaging out these dissimilarities. $D^{inst}$ would preserve these dissimilarities, but it could be difficult for the classifier to select only these relevant dissimilarities due to the high dimensionality of the representation. As an example, consider an image categorization problem, where an image is a bag, and an image region or patch is an instance. If many images in the training set contain regions that include the sky, the dissimilarities to the sky instances in $D^{inst}$ will provide heavily correlated information about the bags. Therefore, $D^{inst}$ could contain many redundant (but not necessarily uninformative) dissimilarities.

On the other hand, when most instances in a bag are informative, we would expect $D^{bag}$ to perform well. $D^{inst}$ would still have access to all the informative dissimilarities, however, selecting a few relevant dissimilarities might not be the best strategy if most instances are, in fact, relevant for the classification problem. The problem of being unable to specify how many dissimilarities are informative, still holds in this case.

\subsection{Classifier and Informative prototypes}

In this work we consider linear classifiers $(\mathbf{w}, w_0)$ such that $f(\mathbf{d}) = \mathbf{w}^{\intercal}\mathbf{d} + w_0$ and $\mathbf{w}$ is an $M$-dimensional vector. The entries of $\mathbf{w}$ correspond to the weights assigned to each of the prototypes, either bags or instances. These weights are found by minimizing an objective function of the form: 

\begin{equation}
\min_{\mathbf{w}} \mathcal{L}(\mathbf{w},\mathcal{T}) ) + \lambda \Omega(\mathbf{w})
\end{equation}

$\mathcal{L}$ is the loss function evaluated on the training set, such as the logistic (for a logistic classifier) or hinge loss (for a support vector classifier or SVM). $\Omega$ is a regularization function of the weight vector and is often the $l_2$ or the $l_1$ norm. The $l_2$ norm typically results in most coefficients of $\mathbf{w}$ being non-zero, while the $l_1$ norm promotes sparsity, i.e., only some of the coefficients have non-zero values. $\lambda$ is a parameter that trades off the loss with the constraints on the weight vector, and therefore influences the final values in $\mathbf{w}$. 

A larger coefficient in $\mathbf{w}$ means the dissimilarity was found to be discriminative by the classifier, we therefore can examine the coefficients to discover which prototypes are more informative. However, the low sample size and high dimensionality/redundancy of feature space can make it difficult to find the $\mathbf{w}$ that leads to the best performance on a held-out test set. 

One way to address the problem of feature redundancy is by a so-called filter approach: evaluating features or subsets of features, prior to training a classifier, and reducing the dimensionality of the training set. However, selecting features individually may disregard important dependencies within the data, and selecting subsets of features is computationally expensive, and may lead to overtraining~\cite{lai2006random}. 

Another alternative is to use a sparse classifier that performs feature selection and classification simultaneously, such as the $l_1$-norm SVM~\cite{zhu20041} or Liknon classifier~\cite{bhattacharyya2003simultaneous}, used in MILES. However, this requires cross-validation to set $\lambda$, which is ill advised in case the training set is small already. A common consequence of such problems is that a poor set of features might be chosen for the testing phase. 

One could argue that clustering the prototypes and selecting the cluster centers as prototypes would alleviate the problem of redundancy in dissimilarities. However, note that redundant dissimilarities are not necessarily caused by similar instances because we are considering minimum distances, not all distances of a set of points. Furthermore, clustering has the risk of excluding informative instances in sparsely populated parts of the feature space, because these would not have any neighbors to form a cluster with. 




\section{Proposed Approach}
\subsection{Random Subspace Ensembles}



We can see the bag and instance representations as two alternative ways of combining dissimilarities of different instances: by averaging or by concatenating.  If we view these approaches as ways of combining different sources of information, a third alternative, ensembles, springs to mind. 

The random subspace method (RSM)~\cite{ho1998random} is one way to create an ensemble that is particularly geared at small sample size, high-dimensional data. Each classifier is built on a lower-dimensional subspace of the original, high-dimensional feature space. This strategy addresses both aspects of a successful ensemble: accurate and diverse classifiers~\cite{kuncheva2003measures,brown2005diversity}. Subsamping the feature space reduces the dimensionality for the individual base classifiers, therefore allowing for more accurate classifiers. Resampling of features introduces diversity~\cite{kuncheva2003measures}, i.e. decorrelates the classifier decisions, which improves the performance of the overall ensemble. 

More formally, the RSM ensemble consists of the following components:

\begin{itemize}

\item Number of subspaces $L$ to be sampled 

\item Numbers of features $\{s_1 \ldots s_L\}$ (or just $s$ if $s_i = s_j \forall i,j$) to be selected for each subspace. 

\item Base classifier $f$, which is applied to each subspace. We denote the trained classifiers by $\{f_1, \ldots, f_L\}$. 

\item Combining function $g$, which for a test feature vector $\mathbf{d}$, combines the outputs into a final classifier $F(\mathbf{d}) = g(f_1(\mathbf{d}), \ldots f_L(\mathbf{d}))$.  
\end{itemize}

RSM is interesting in high-dimensional problems with high feature redundancy~\cite{skurichina2001bagging}. For example, the expression levels of co-regulated (both influenced by another process) genes will provide correlated information about whether a subject has diabetes or not. Other genes may be irrelevant to diabetes, only adding noise. We typically do not have prior knowledge about the number of underlying processes that are responsible to diabetes, i.e., the amount of redundancy is unknown. This increases the number of possible relevant feature sets, and makes selecting only the relevant features more difficult. RSM decreases this risk, simplifying the feature selection problem for each individual classifier, and by still allowing access to all the (possibly relevant) features, thus letting the classifiers correct each other. Other examples where RSM is a successful approach include functional magnetic resonance imaging (fMRI) data~\cite{kuncheva2010random}, microarray data~\cite{bertoni2005bio} and hyperspectral data~\cite{ham2005investigation}.



The different prototypes in MIL may also provide redundant information, but we do not know in advance how many such redundant prototypes there might be. Furthermore, many MIL problems are small sample size problems in the number of bags, so additional classifier evaluations during training are undesirable. Therefore we believe that RSM can be an attractive method to address dissimilarity-based MIL, and to combine the strengths of the dissimilarity space with those of ensemble classifiers.


\subsection{Choice of Subspaces}

There are two alternatives for how the subspace classifiers can be defined, which we first introduced in~\cite{cheplygina2013combining}:

\begin{itemize}
\item By choosing each prototype bag as a subspace, i.e. the subspace is formed by the dissimilarities to the instances of a prototype bag.  We denote this representation by $D^{BS}$, where $BS$ stands for Bag Subspaces. The RSM parameters are straightforward here: $L=M$ and and the subspace dimensionalities $s_i$ correspond to the bag sizes $n_i$.

\item By choosing each subspace randomly. We denote this representation by $D^{RS}$, where $RS$ stands for Random Subspaces. $D^{RS}$ offers more flexibility with regard to the RSM parameters. In~\cite{cheplygina2013combining}, we used default parameters $L=M$ and $s = \frac{1}{N}\sum_i n_i$. However, alternative settings are possible as well, and we will demonstrate further on in this paper that other choices (which can be set by rules of thumb rather than cross-validation), can in fact improve the results significantly. 
\end{itemize}

Note that these alternatives are both slightly different from RSM because the dissimilarity representation depends on the training set. In traditional RSM, all features are available to the classifier at any split of the training and test data, whereas with $D^{BS}$ and $D^{RS}$, the features are defined through dissimilarities to the training set, which obviously changes with every training-test split. However, we still expect there to be a relationship between how RSM parameters, and the choices in $D^{BS}$ and $D^{RS}$, affect ensemble performance. 

We provide a summary of the ensembles, as well as the single classifier dissimilarity representations in Table~\ref{tab:rep}.

\begin{table}
\centering
\begin{tabular}{ccc}
Representation & Dimensionality & Classifiers \\
\hline
$D^{bag}$ & $M$  & 1 \\
$D^{inst}$ & $\sum_i n_i $ & 1 \\
$D^{BS}$ & $\{n_1,\ldots n_M\}$ & $M$ \\
$D^{RS}$ & any & any \\
\end{tabular}
\caption[]{Different ways for constructing dissimilarity representations. $D^{bag}$ consists of dissimilarities to bags in the training set (one for each bag), whereas $D^{inst}$ consists of dissimilarities to instances in the training set. In $D^{BS}$, a separate classifier is built on each prototype's instance dissimilarities. In $D^{RS}$, classifiers are built on random selections of all available instances.}
\label{tab:rep}
\end{table}

\subsection{Illustrative example}
\begin{figure}[ht!]
 \centering
 \subfloat[]{
  \includegraphics[width=0.90\columnwidth]{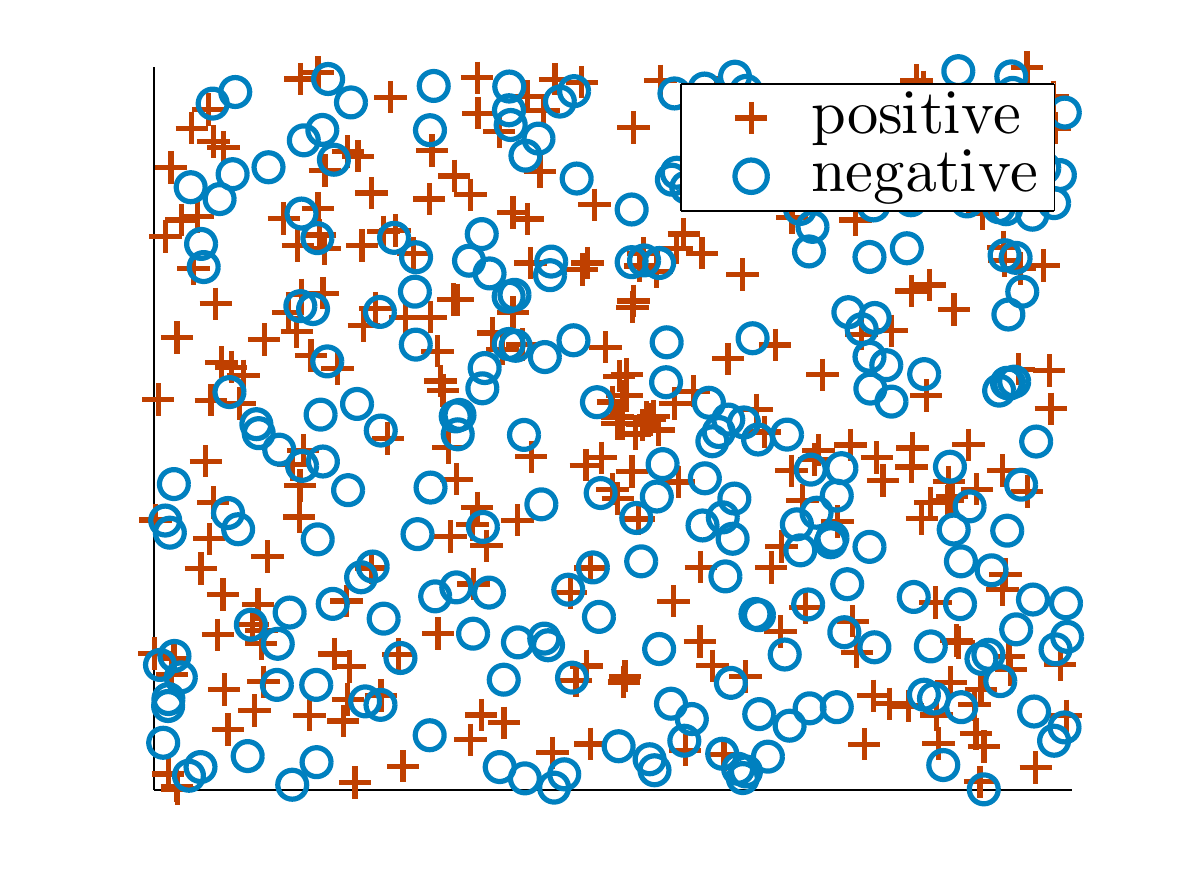}
   }
   
 \subfloat[]{
  \includegraphics[width=0.90\columnwidth]{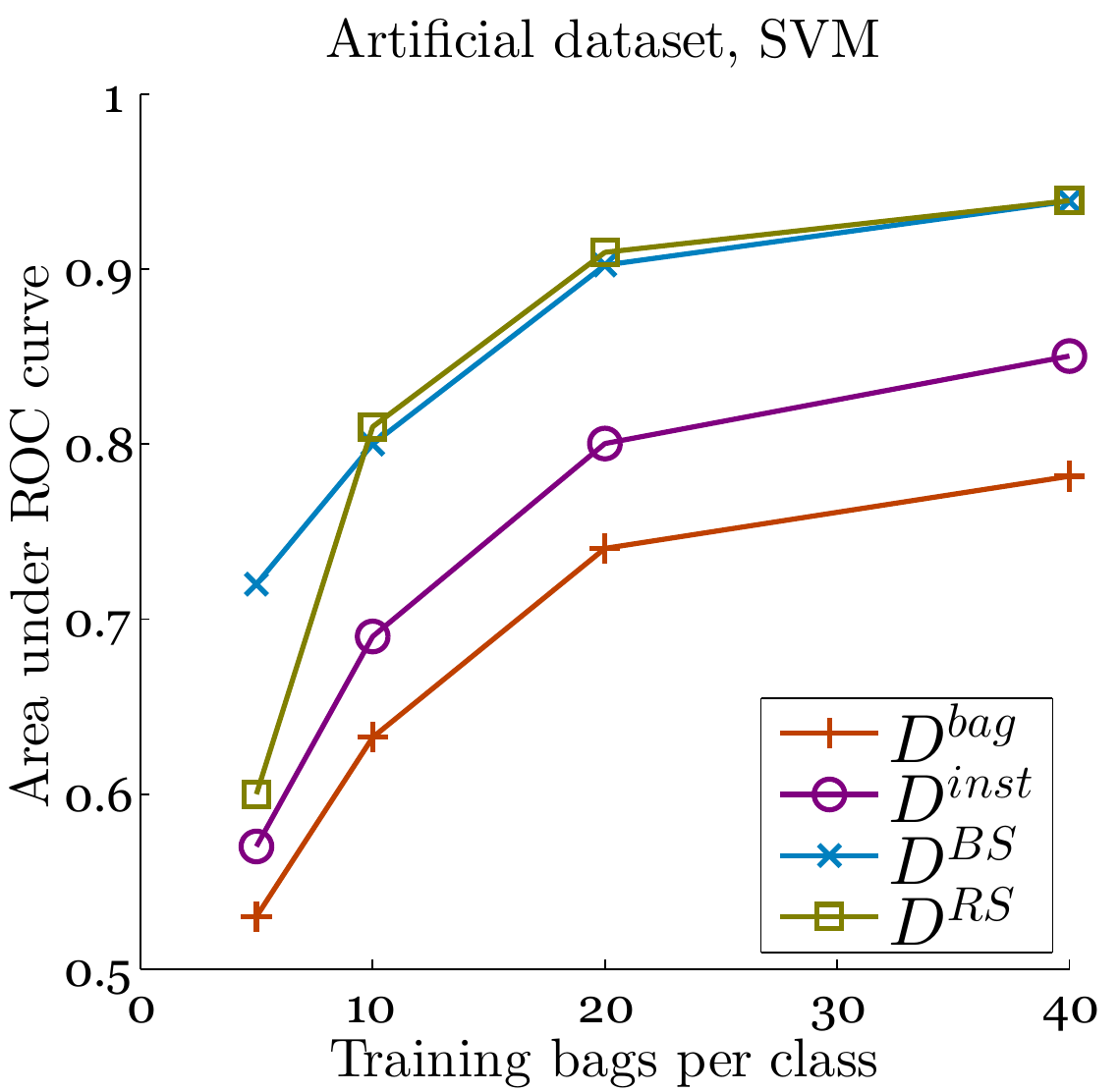}
   }
           
 \caption{Top: Artificial 2D MIL problem with informative instances in the center. Bottom: Learning curves for dissimilarity-based classifiers on this dataset. The amount of uninformative, and redundant instances deteriorates performances of $D^{bag}$ and $D^{inst}$, but the ensemble methods $D^{BS}$ and $D^{RS}$ are more robust against these problems.}
 \label{fig:concept}
\end{figure}

The basic intuition about the benefit of the proposed ensembles is illustrated by the artificial problem in the top of Fig.~\ref{fig:concept}. This is the classical MIL problem from~\cite{maron1998framework}. This dataset contains bags with 50 two-dimensional instances. The instances from the bags are uniformly distributed in a square, and the positive bags contain at least one feature vector from a concept region that is located in the middle of the square. Only the dissimilarities of the concept instances are informative. Averaging over the dissimilarities as in $D^{bag}$ dilutes these informative features, and indeed, the learning curves in the bottom of Fig.~\ref{fig:concept} show that $D^{bag}$ performs poorly here. $D^{inst}$ has trouble selecting only the informative dissimilarities because many dissimilarities are uninformative, and because dissimilarities of the informative instances are correlated. The ensemble methods are more robust against these problems and achieve the best performances (Fig.~\ref{fig:concept}, bottom). 

\section{Experiments}

\subsection{Data and Setup}

We provide a list of datasets we used in Table~\ref{tab:datasets}. The Musk problems~\cite{dietterich1997solving} are traditional benchmark MIL problems about molecule activity, Mutagenesis~\cite{srinivasan1995comparing} is a drug activity prediction problem. Fox, Tiger and Elephant~\cite{andrews2002multiple} are benchmark image datasets. African and Beach are also image datasets originating from a multi-class scene recognition problem~\cite{chen2006miles}, but here formulated as one-against-all problems. The dataset alt.atheism originates from the Newsgroups data~\cite{zhou2009multi}, and are concerned with text categorization. Brown Creeper and Winter Wren are both bird song~\cite{briggs2012acoustic} datasets, where the goal is to classify whether a particular bird species can be heard in an audio recording.

We preprocess the data by scaling each feature to zero mean and unit variance. The scaled data is used to compute the dissimilarity representations. The instance dissimilarity function is defined as the squared Euclidean distance: $d(\mathbf{x}_i,\mathbf{x}_j) = (\mathbf{x}_i-\mathbf{x}_j)^{\intercal}(\mathbf{x}_i-\mathbf{x}_j)$. 

For the base classifier $f$, we consider linear classifiers as described in Section 2. We used several linear classifiers: logistic, 1-norm SVM and a linear SVM, where the primal formulation is optimized~\cite{chapelle2007training}. The trade-off parameter $\lambda$ is set to 1 by default. The common characteristic of these classifiers is that we can inspect the weight vector $\mathbf{w}$ to determine which dissimilarities are deemed to be more important by the classifier. Although the individual performances of the classifiers differ, we observe similar trends (such as relative performances of two different ensembles) for these choices. We therefore only show results for the linear SVM. 


For the combining function $g$, we average the posterior probabilities, which are obtained after normalizing the outputs of each classifier to the [0, 1] range. We also considered majority voting, and using the product and the maximum of the posterior probabilities, but overall, averaging led to the best performances. Furthermore, averaging posterior probabilities performs well in practice in other problems as well~\cite{tax2000combining,cheplygina2011pruned}.

The metric used for comparisons is area under the receiver-operating characteristic (AUC)~\cite{bradley1997use}. This measure has been shown to be more discriminative than accuracy in classifier comparisons~\cite{huang2005using}, and more suitable for MIL problems~\cite{tax2008learning}.

\begin{table}[h]
\centering

\begin{tabular}{|l |  p{0.7cm} p{0.7cm} p{0.7cm} p{0.7cm} |}
\hline
Dataset & +bags & -bags & total & average  \\
\hline
Musk 1  & 47 & 45 & 476 & 5   \\ 
Musk 2 & 39 & 63 & 6598 & 65  \\
Fox  & 100 & 100 & 1302 & 7  \\
Tiger & 100 & 100 & 1220  & 6  \\
Elephant  & 100 & 100 & 1391  & 7 \\
Mutagenesis 1 & 125 & 63 & 10486 & 56  \\
African & 100 & 1900 & 7947 & 8 \\
Beach& 100 & 1900 & 7947 & 8 \\
Alt.atheism  & 50 & 50 & 5443  & 54   \\
Brown Creeper & 197 & 351 & 10232  & 19  \\
Winter Wren  & 109 & 439 & 10232 & 19  \\
\hline
\end{tabular}
\caption{MIL datasets, number of bags, instances, the average number of instances per bag. The datasets are available online at \texttt{http://www.miproblems.org}}
\label{tab:datasets}
\end{table}

\subsection{Subspace Experiments}

\begin{figure*}[ht!]
 \centering
 \subfloat[]{
  \includegraphics[width=0.40\textwidth]{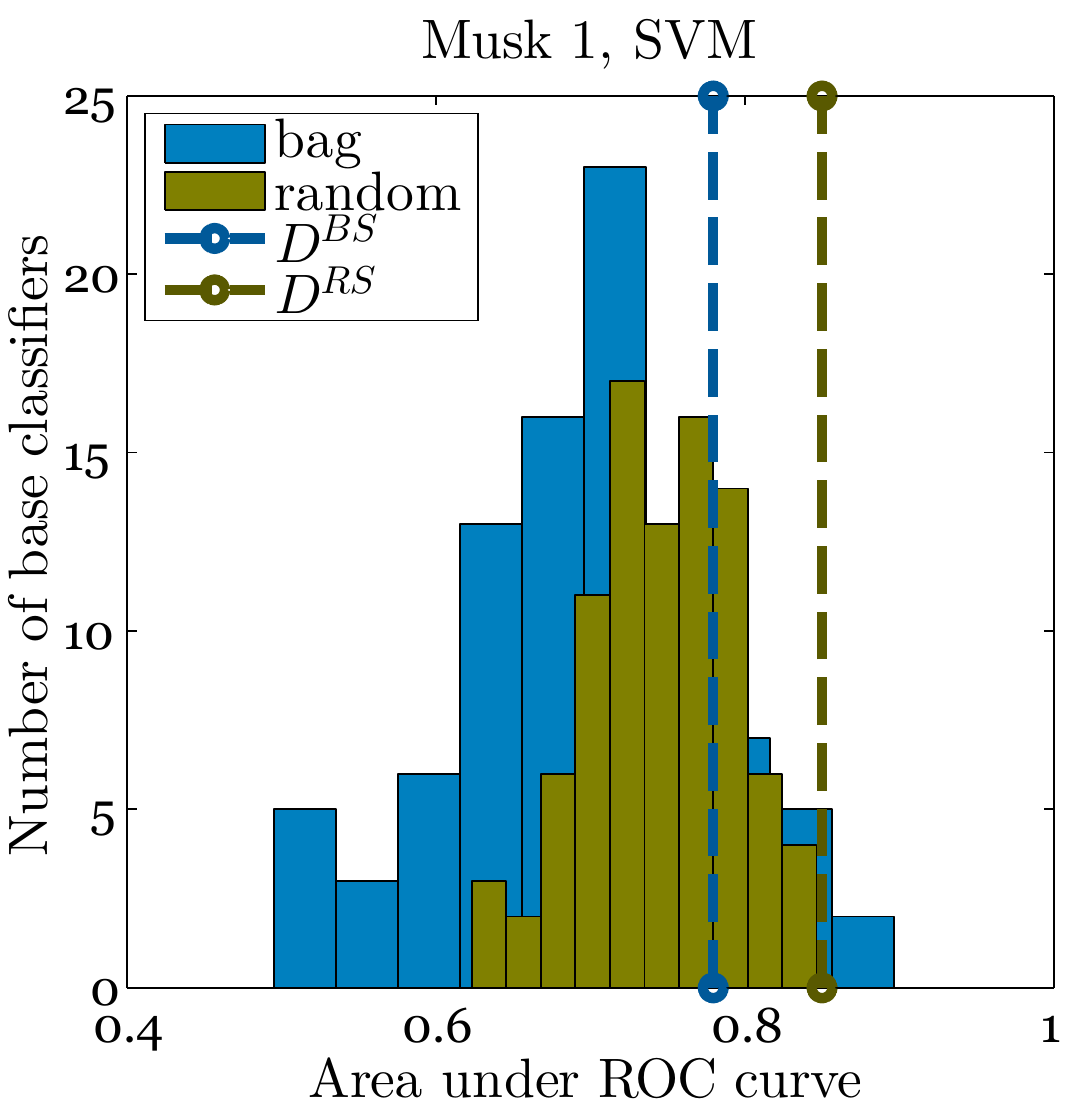}
   }
 \subfloat[]{
  \includegraphics[width=0.40\textwidth]{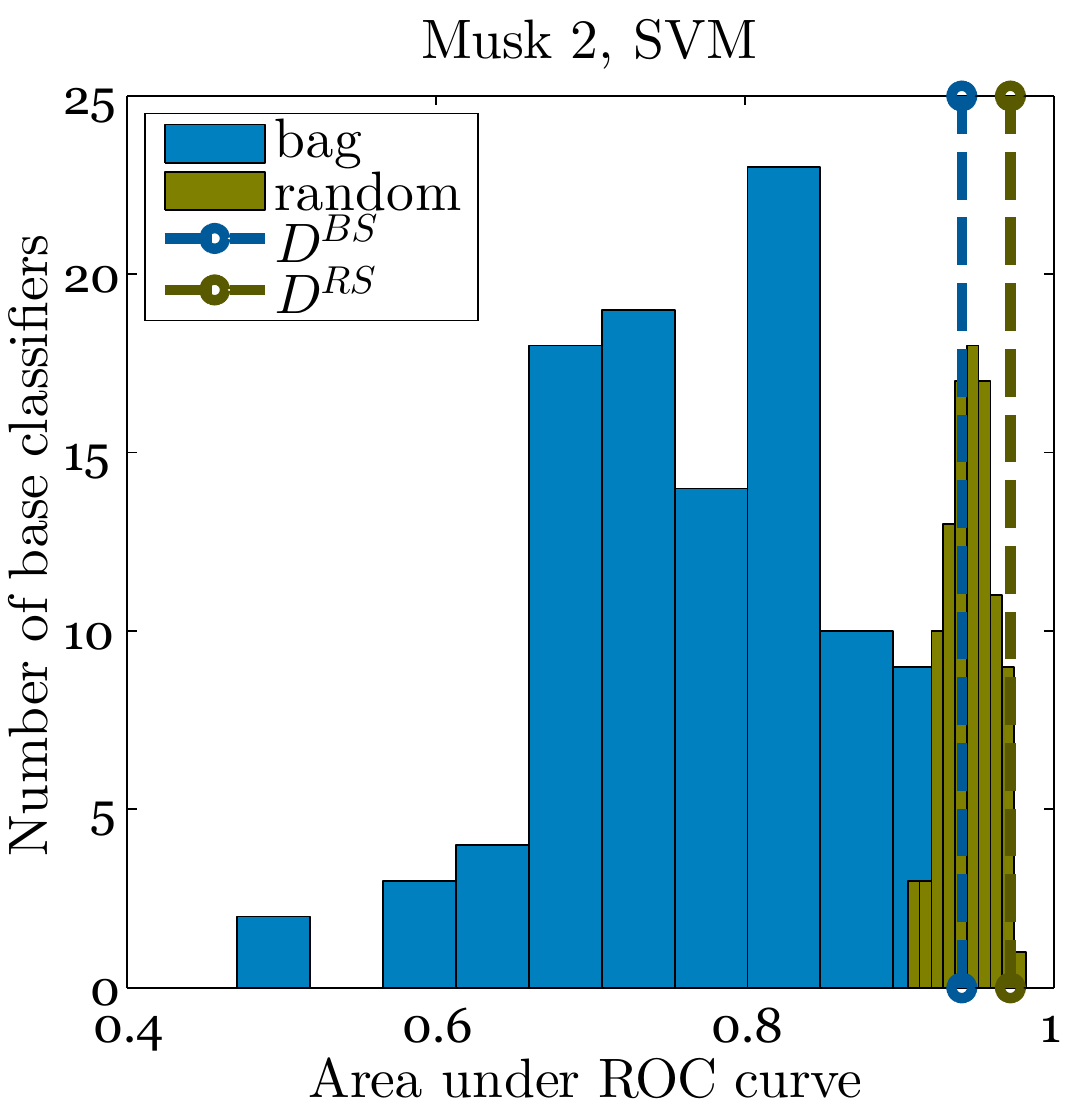}
   }
   
\subfloat[]{
  \includegraphics[width=0.40\textwidth]{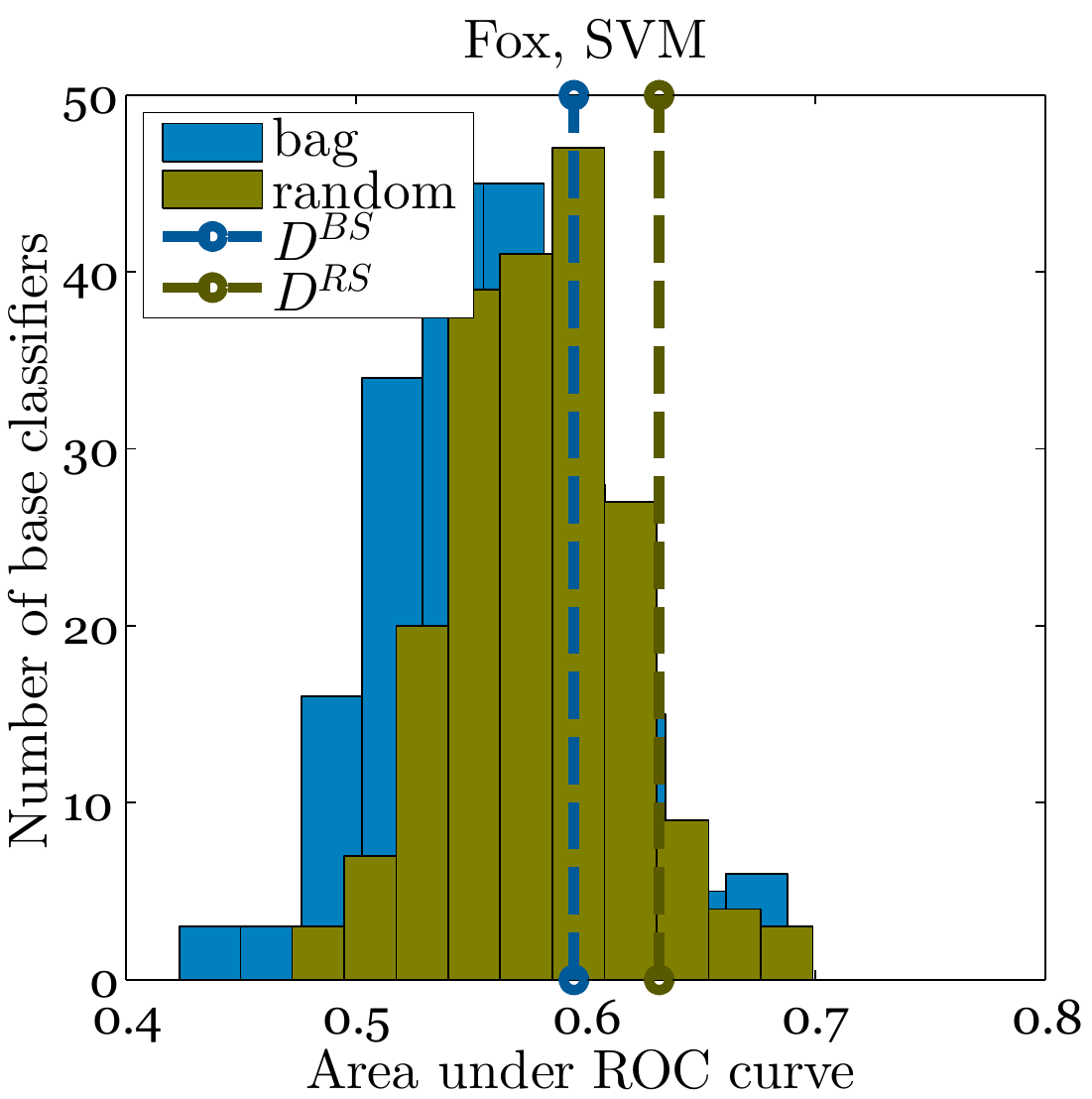}
   }
 \subfloat[]{
  \includegraphics[width=0.40\textwidth]{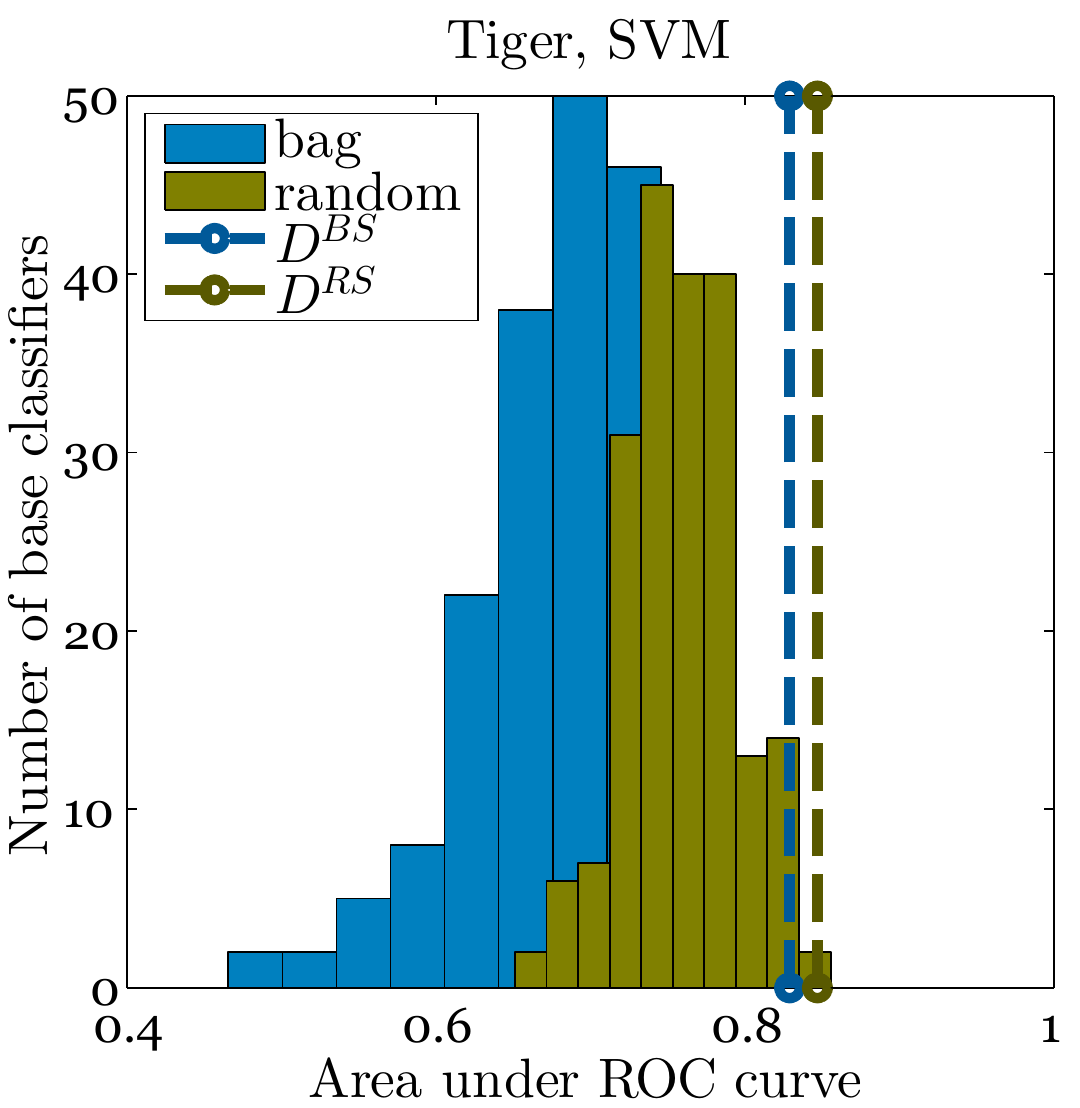}
   }
   
   \subfloat[]{
  \includegraphics[width=0.40\textwidth]{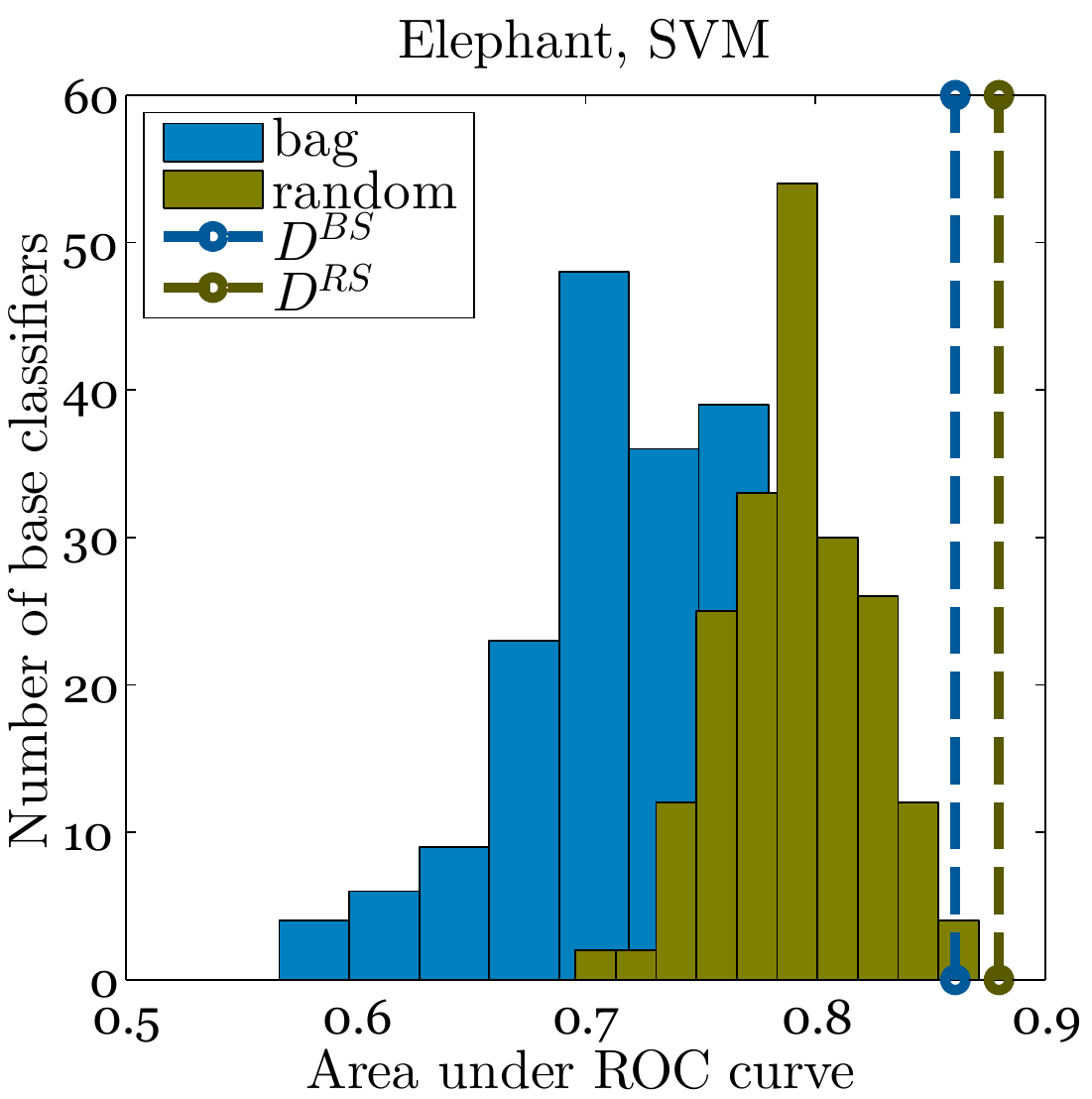}
   }
 \subfloat[]{
  \includegraphics[width=0.40\textwidth]{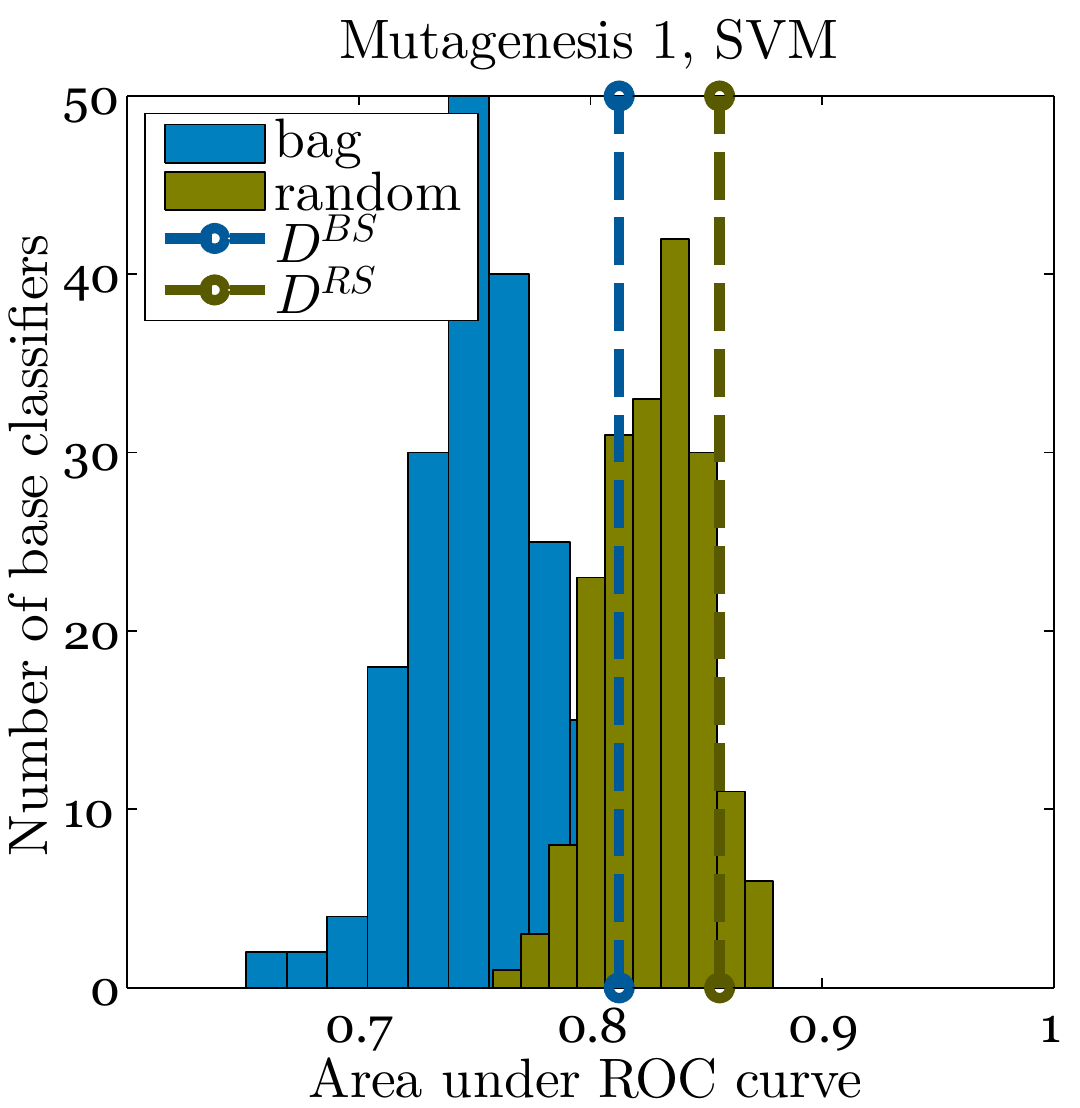}
   }
       
 \caption[]{Distributions of AUC performances of individual bag subspace classifiers}
 \label{fig:hists}
\end{figure*}

We start by comparing the two alternatives of creating the subspaces, $D^{BS}$ and $D^{RS}$. For simplicity, we base the parameters of $D^{RS}$ on those for $D^{BS}$, as in~\cite{cheplygina2013combining}: $M$ subspaces, each subspace with dimensionality $\frac{1}{N} \sum_i n_i$. We use a linear SVM as the classifier ($C$ parameter is set to 1 by default) and perform 10-fold cross-validation. Fig.~\ref{fig:hists} shows the distributions of the individual classifier performances and the ensemble performance for both representations.

The results show that overall,
\begin{enumerate}
\item the random subspaces are more informative than bag subspaces, and 
\item the random subspace ensemble is better at improving upon the base classifiers.
\end{enumerate}

\begin{figure}[ht!]
 \centering
 \subfloat[]{
  \includegraphics[width=0.90\columnwidth]{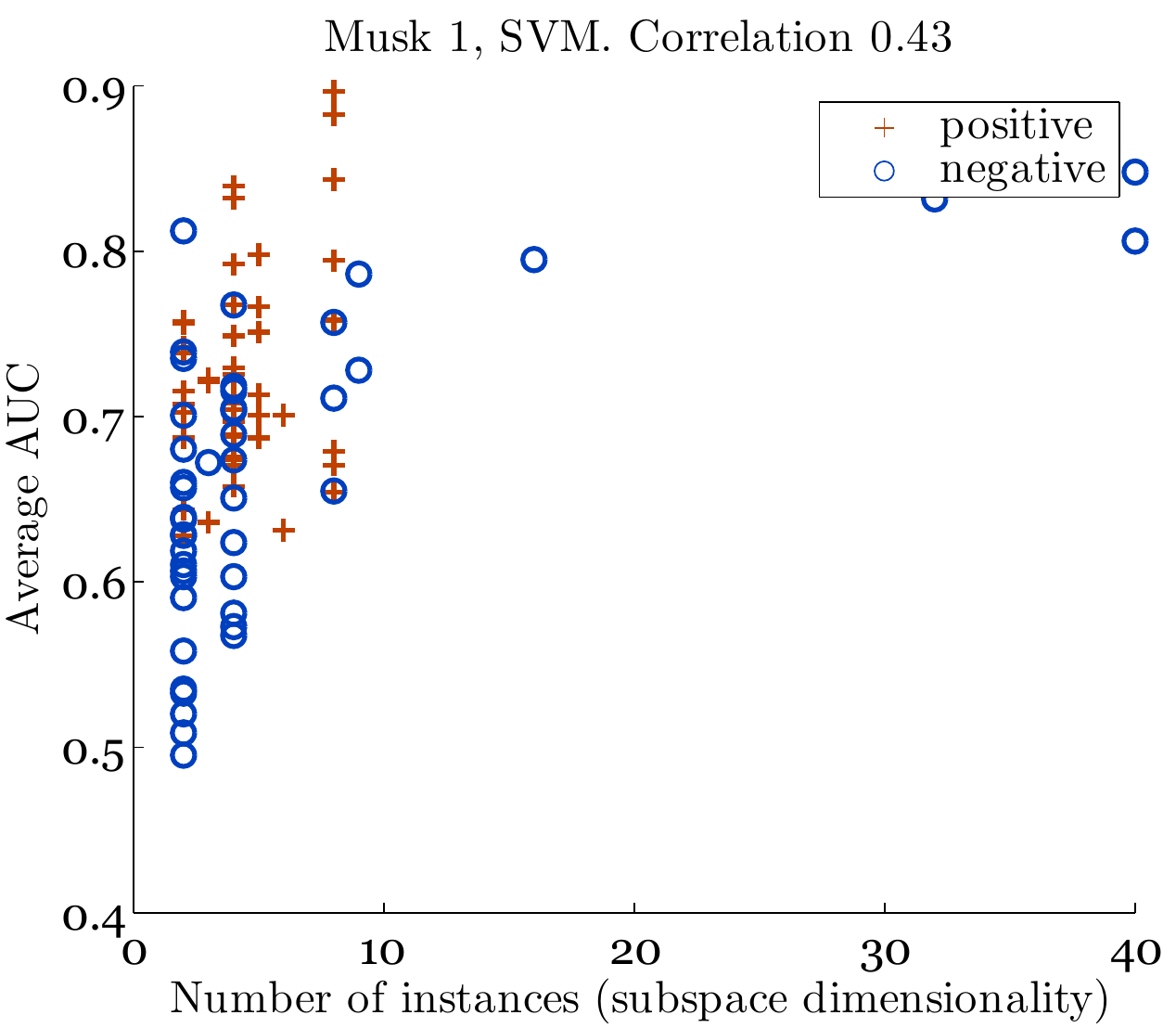}
   }
   
 \subfloat[]{
  \includegraphics[width=0.90\columnwidth]{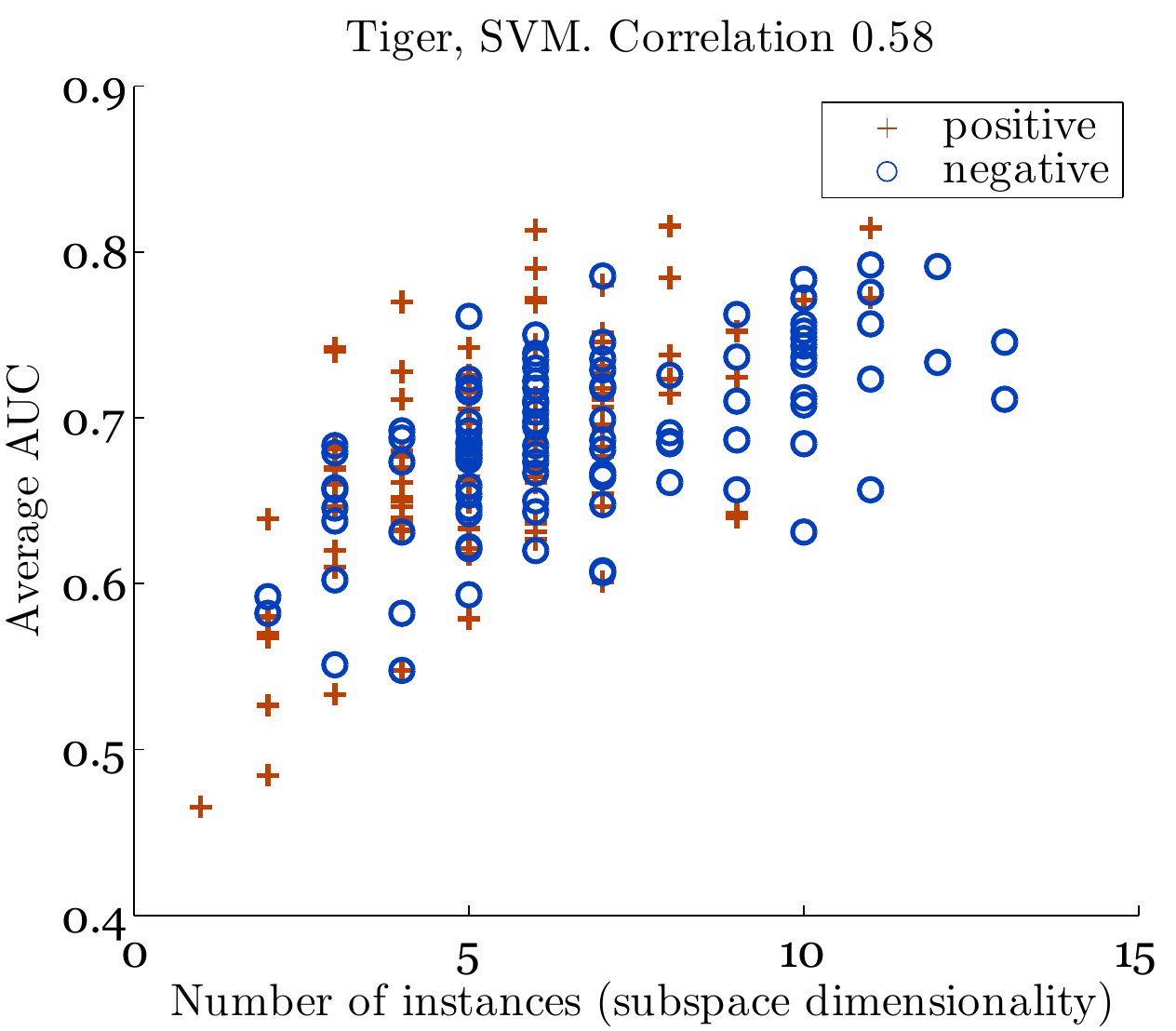}
   }
   
 \caption[]{Relationship of bag size, bag label, and AUC performance of the dissimilarity subspace formed by that bag}
 \label{fig:scatterbags}
\end{figure}

Why do the random subspaces perform better than the bag subspaces? One difference might be the subspace size: the bag subspaces have variable dimensionalities, whereas the random subspaces are equally large. For $D^{BS}$, we plot the bag size against the performance of that subspace. A few of the results are shown in Fig.~\ref{fig:scatterbags}. In all datasets except alt.atheism, we find medium to high correlations between these quantities. Therefore, as prototypes, small bags are not very informative. This might seem counterintuitive in a MIL problem, because small bags are less ambiguous with respect to the instance labels under the standard assumption. The fact that large, ambiguous bags are better at explaining the class differences suggests that most instances are informative as prototypes.

The informativeness of most instances as prototypes is supported by the relationship between the bag label and the subspace performance in the plots. Although for a fixed bag size, positive bag subspaces perform better on average, negative bags can also be very good prototypes. This is also true for random bags, for which we do not have labels, but for which we can examine the percentage of instances, which were sampled from positive bags. We found no correlations between the "positiveness" of the random subspaces and their performance.
This provides opportunities for using unlabeled data in a semi-supervised way: unlabeled bags can be used to extend the
dissimilarity representation to improve performance, similar to the approach in~\cite{dinh2012study}.

\begin{figure}[ht!]
 \centering
 \subfloat[]{
  \includegraphics[width=0.90\columnwidth]{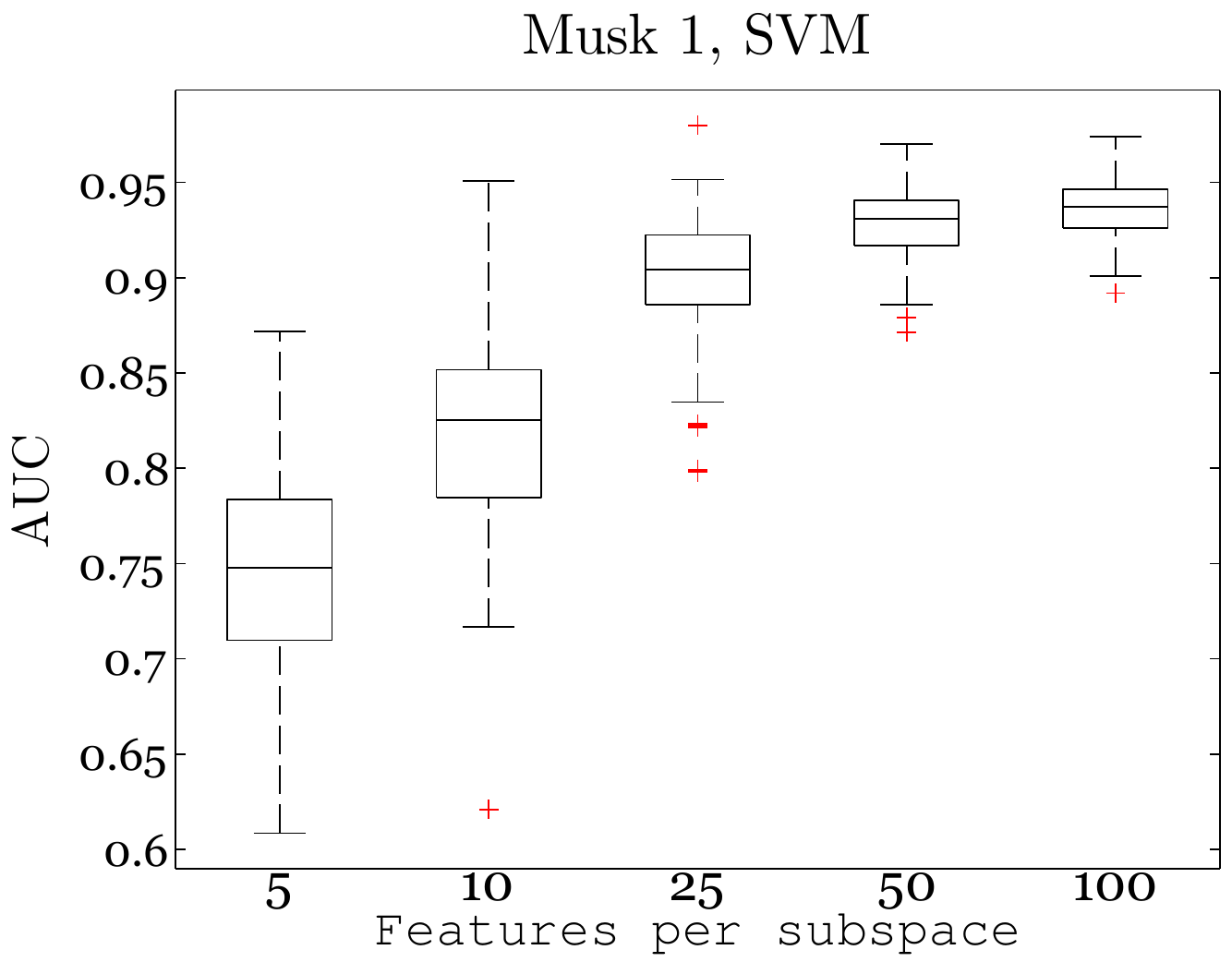}
   }
   
 \subfloat[]{
  \includegraphics[width=0.90\columnwidth]{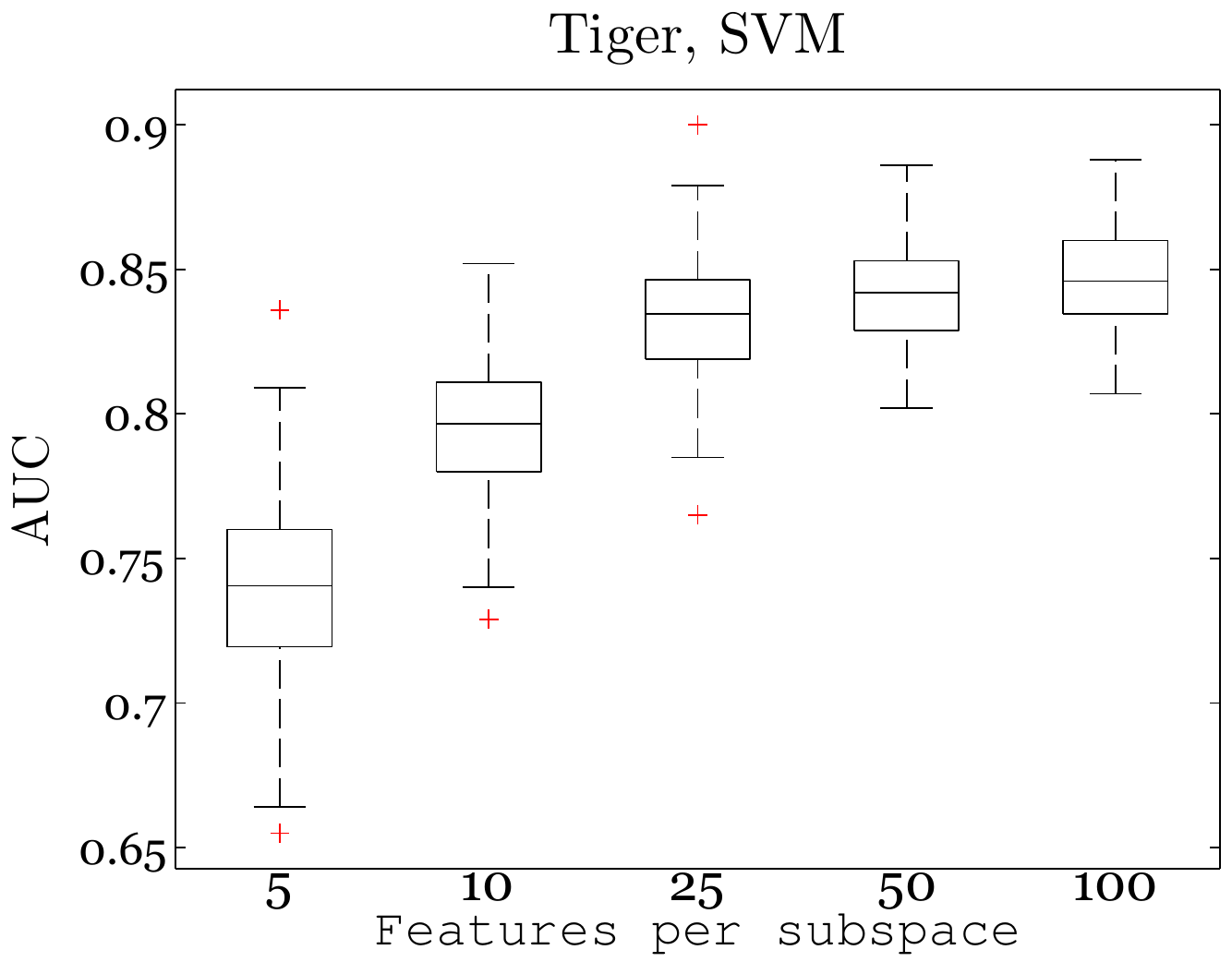}
   }

 \caption[]{Distributions of AUC performances of individual subspace classifiers, for different dimensionalities of the subspaces.}
 \label{fig:res_scattersize1}
\end{figure}

The results with respect to the bag size suggest that it is advantageous to combine classifiers built on higher-dimensional subspaces. We therefore investigate the effects of subspace dimensionality in $D^{RS}$, where this parameter can be varied. We vary the subspace size for each classifier from 5 to 100 features, which for most datasets, would be larger than the default dimensionalities used previously, as shown in Table~\ref{tab:datasets}. We generate 100 subspaces of each dimensionality, and train a linear SVM on each subspace. The classifiers are then evaluated individually. Some typical distributions of performances per subspace dimensionality are shown in Fig.~\ref{fig:res_scattersize1}. For most datasets (except Newsgroups, as will be explained later), larger subspaces lead to more accurate classifiers.

Why is $D^{RS}$ better than $D^{BS}$ at improving upon the individual classifiers? A possible explanation is that the classifiers created by $D^{RS}$ are more diverse than the classifiers of $D^{BS}$. For each set of classifiers, we examine their $L\times L$ disagreement matrix $C$, where each entry $C_{i,j}$ corresponds to the number of test bags for which the $i$-th and $j$-th classifier provide different labels. $C_{i,j}$ can be seen as the distance of the two classifiers. We perform multi-dimensional scaling with each of these distance matrices and map the classifiers into a 2-dimensional classifier projection space~\cite{pkekalska2002discussion}. 

\begin{figure}[ht!]
 \centering

\subfloat[]{
  \includegraphics[width=0.90\columnwidth]{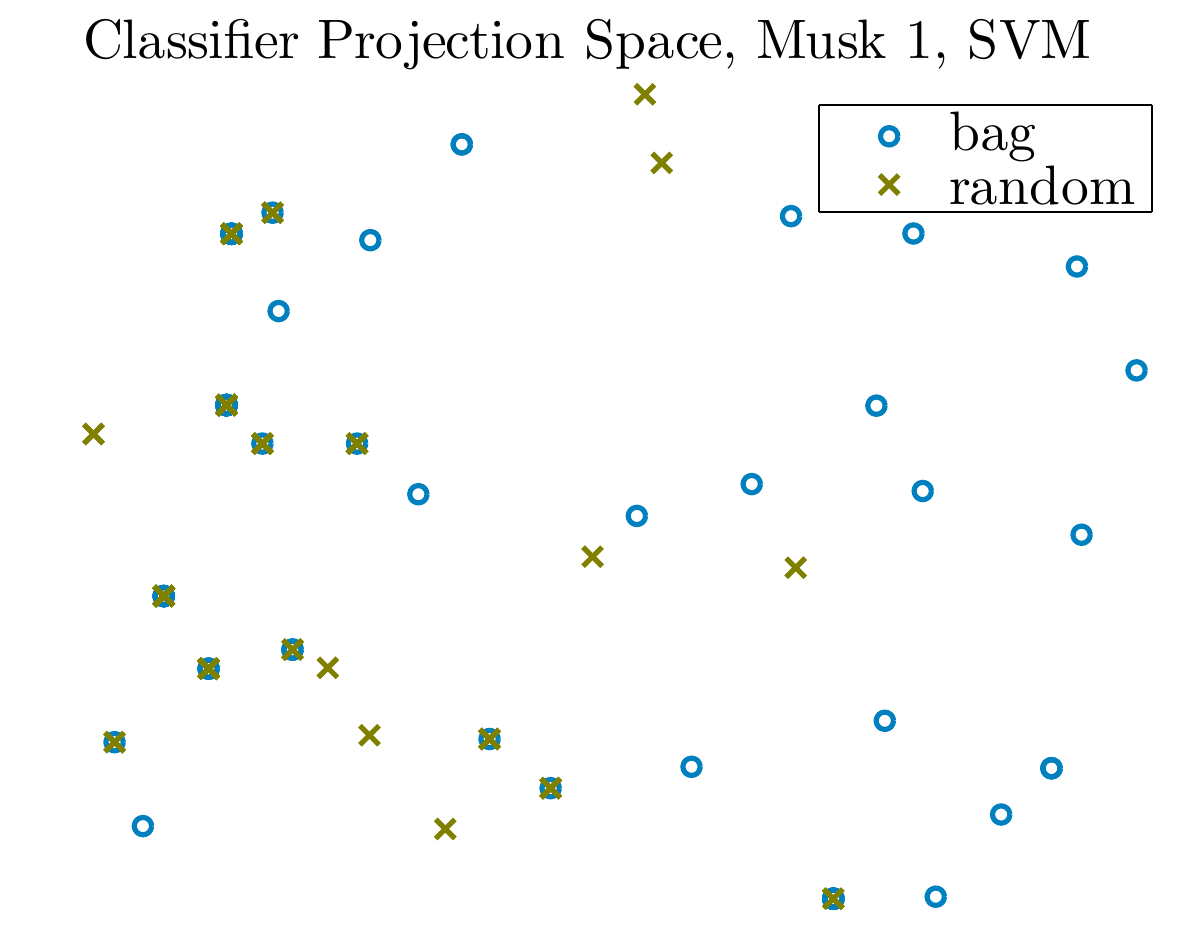}
   }
   
 \subfloat[]{
  \includegraphics[width=0.90\columnwidth]{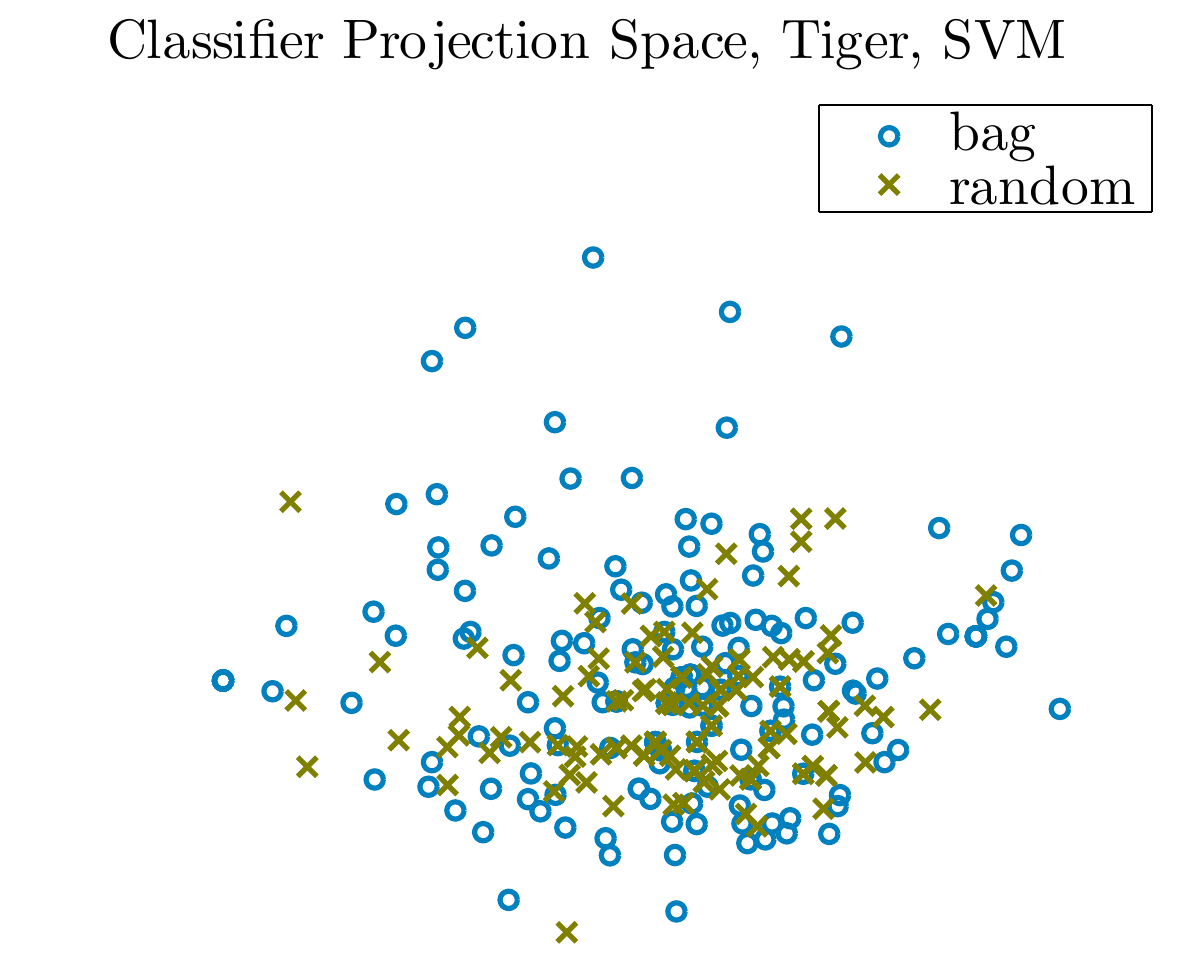}
   }

 \caption[]{Classifier projection spaces. The plots show the relative disagreement of trained subspace classifiers on a test set. The higher the disagreement of two classifiers on the labels of the test set, the larger their distance in the plot. }
 \label{fig:cps}
\end{figure}

The classifier projection spaces for two datasets are shown in Fig.~\ref{fig:cps}. These results are surprising, because the classifiers
in $D^{BS}$ actually cover a larger part of the space, and are therefore more diverse. This diversity alone, however, is not able to improve the overall performance of the ensemble. A possible explanation is that here we are dealing with "bad diversity"~\cite{brown2010good}. For example, a classifier that is 100\% wrong is very diverse with respect to a classifier that is 100\% correct, but not beneficial when added to an ensemble. We showed in Fig.~\ref{fig:scatterbags} that in $D^{BS}$, small bags often result in inaccurate classifiers, which are indeed responsible for higher diversity, but worsen the ensemble performance.


In the experiments, the Newsgroups data shows very different behavior from the other datasets. Most of the subspaces (both bag and random) have nearly random performance, and the performances are not correlated with the bag label or subspace dimensionality.  A possible explanation is that in this dataset, many of the dissimilarities only contain noise, and the informativeness is distributed only across a few dissimilarities. RSM is particularly suitable for problems where the informativeness is spread out over many (redundant) features~\cite{skurichina2001bagging}, which could explain the worse than random performance. Indeed, examining the individual informativeness (as measured by the nearest neighbor error) of each individual dissimilarity, it turns out that more than half of the dissimilarities in alt.atheism have worse than random performance, as opposed to only around 10\% of dissimilarities for the other datasets. 

\begin{figure}[ht!]
 \centering
 \subfloat[]{
  \includegraphics[width=0.90\columnwidth]{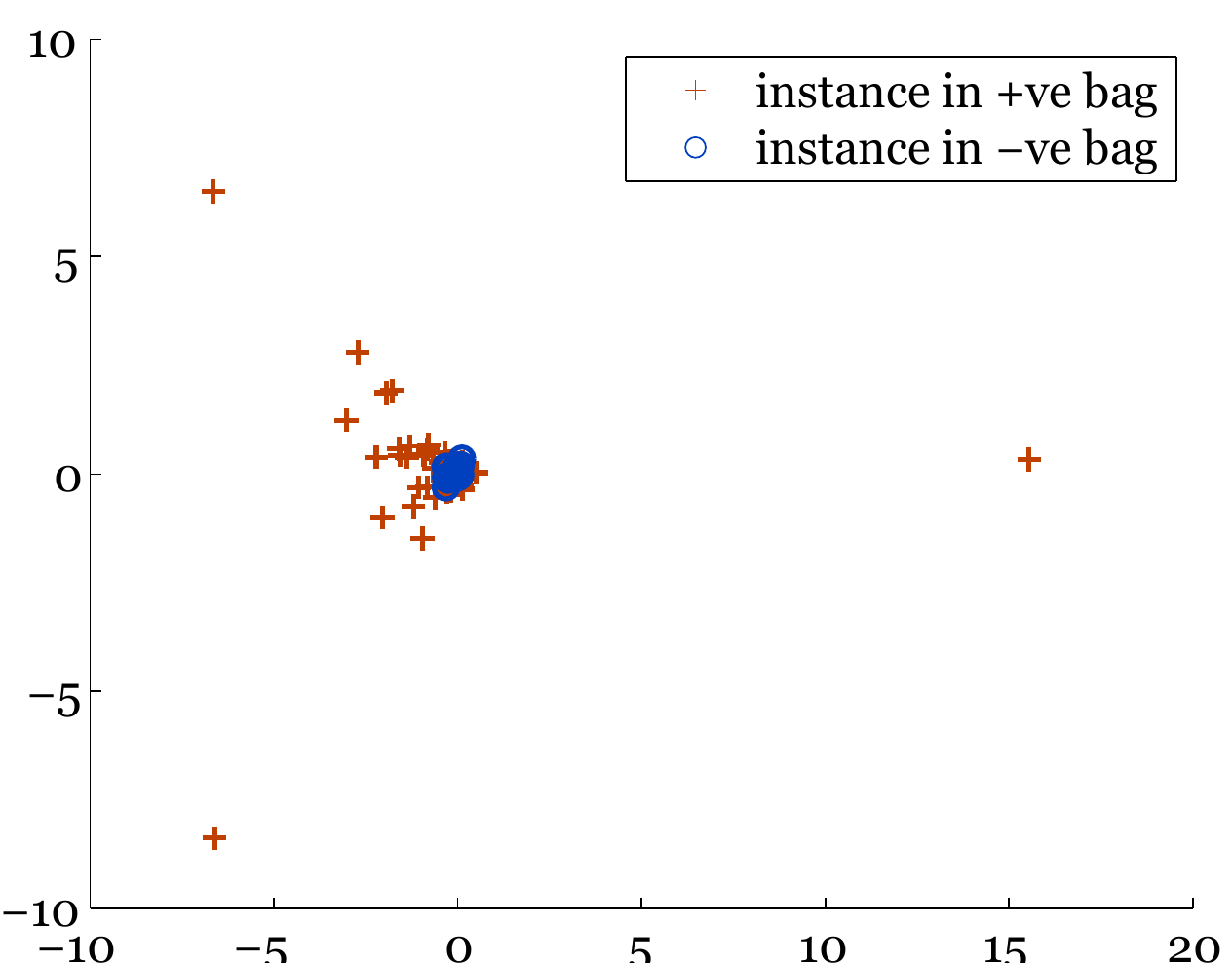}
   }
 \caption[]{Multi-dimensional scaling of the instances in the alt.atheism dataset}
 \label{fig:mds164}
\end{figure}

We have noticed previously~\cite{cheplygina2012class} that positive bags in the Newsgroups data consist of dense cluster of instances and a few "outliers", while negative bags of the dense cluster of instances, and a few outliers. This distribution is caused by the bag of words representation of the data --- while the outlier instances are in fact all positive for the alt.atheism topic, they do not consist of the same words, and are far from each other in the feature space. This situation is shown in Fig.\ref{fig:mds164}. The presence or absence of such outliers is very informative for the bag class. The definition of dissimilarity in (\ref{eq:rep_inst}), however, does not succeed in extracting this information: for any training bag, the instance closest to the prototype outlier instance, is in the dense cluster. In this case the minimum function in the dissimilarity is not suitable for the problem at hand, and much better performances can be obtained by, for instance, using bags and prototypes, and considering the asymmetry of $D^{bag}$~\cite{cheplygina2012class}.


\subsection{Ensemble Experiments}

With several interesting results for the individual performances of the subspace classifiers, we now investigate how the ensemble behaves when both the parameters of interest are varied. The ensembles are built in such a way that classifiers are only added (not replaced) as the ensemble size increases, i.e. an ensemble with 100 classifiers has the same classifiers as the ensemble with 50 classifiers, and 50 additional ones.  

\begin{figure*}[ht!]
 \centering

\subfloat[]{
  \includegraphics[width=0.45\textwidth]{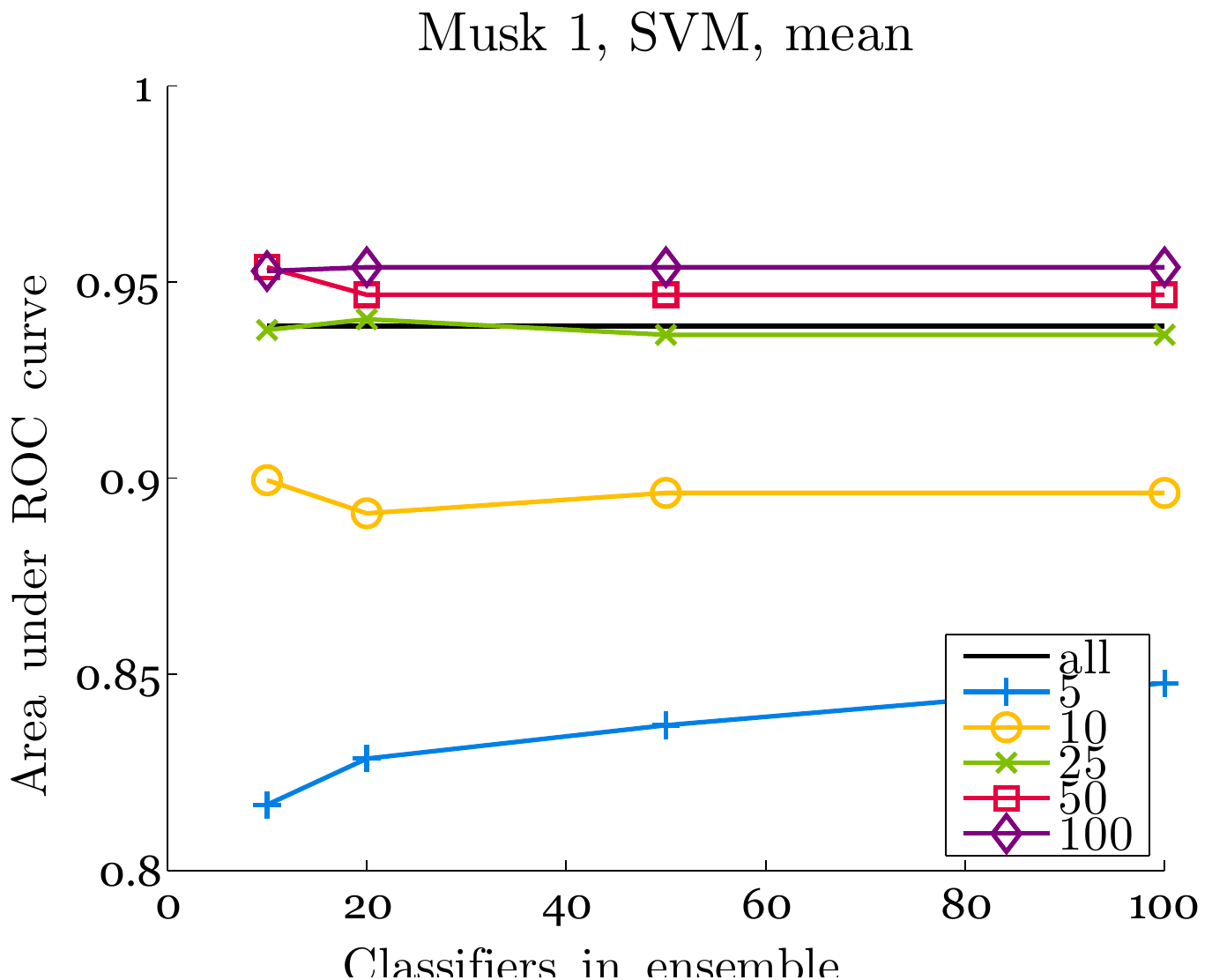}
   }
 \subfloat[]{
  \includegraphics[width=0.45\textwidth]{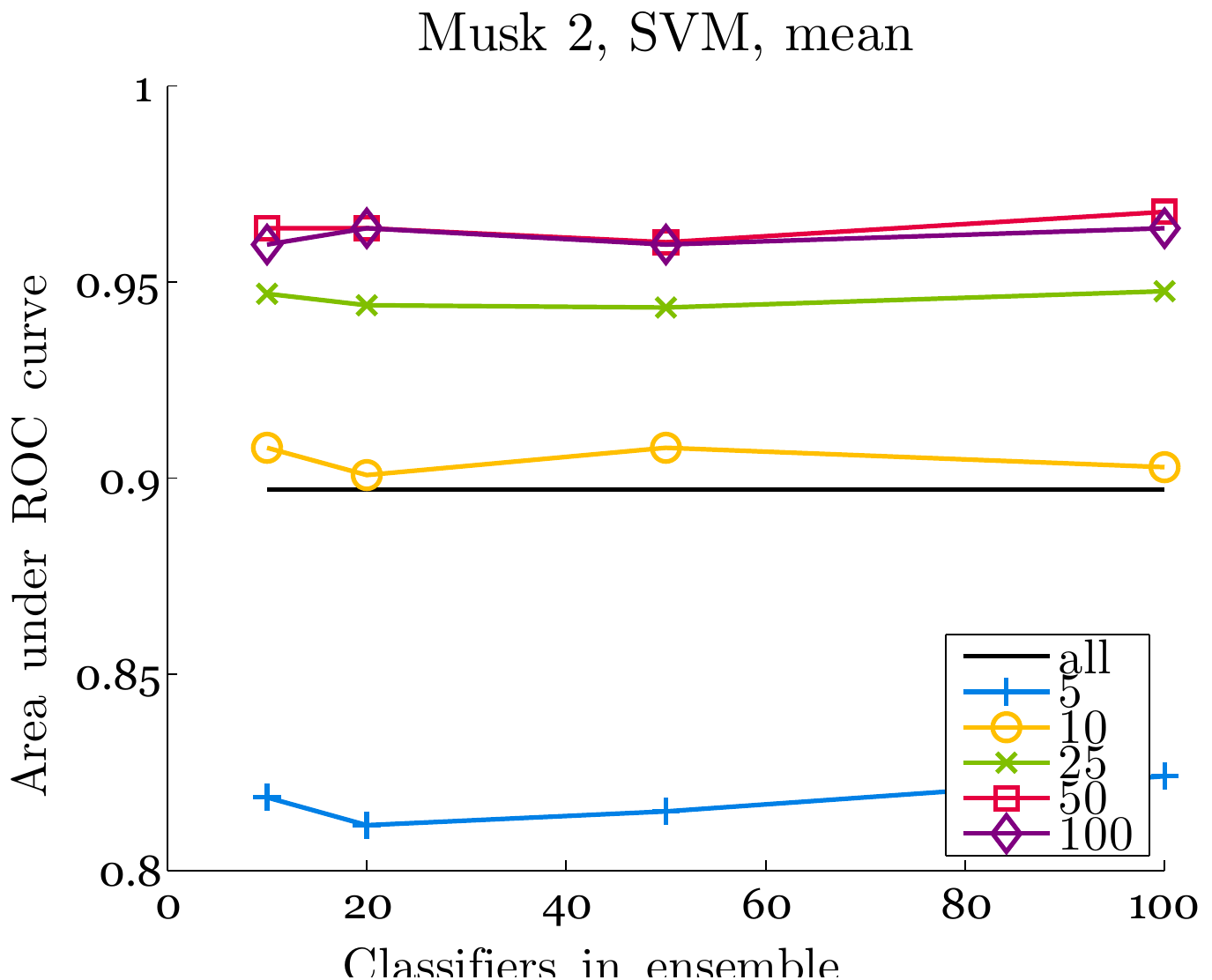}
   }

\subfloat[]{
  \includegraphics[width=0.45\textwidth]{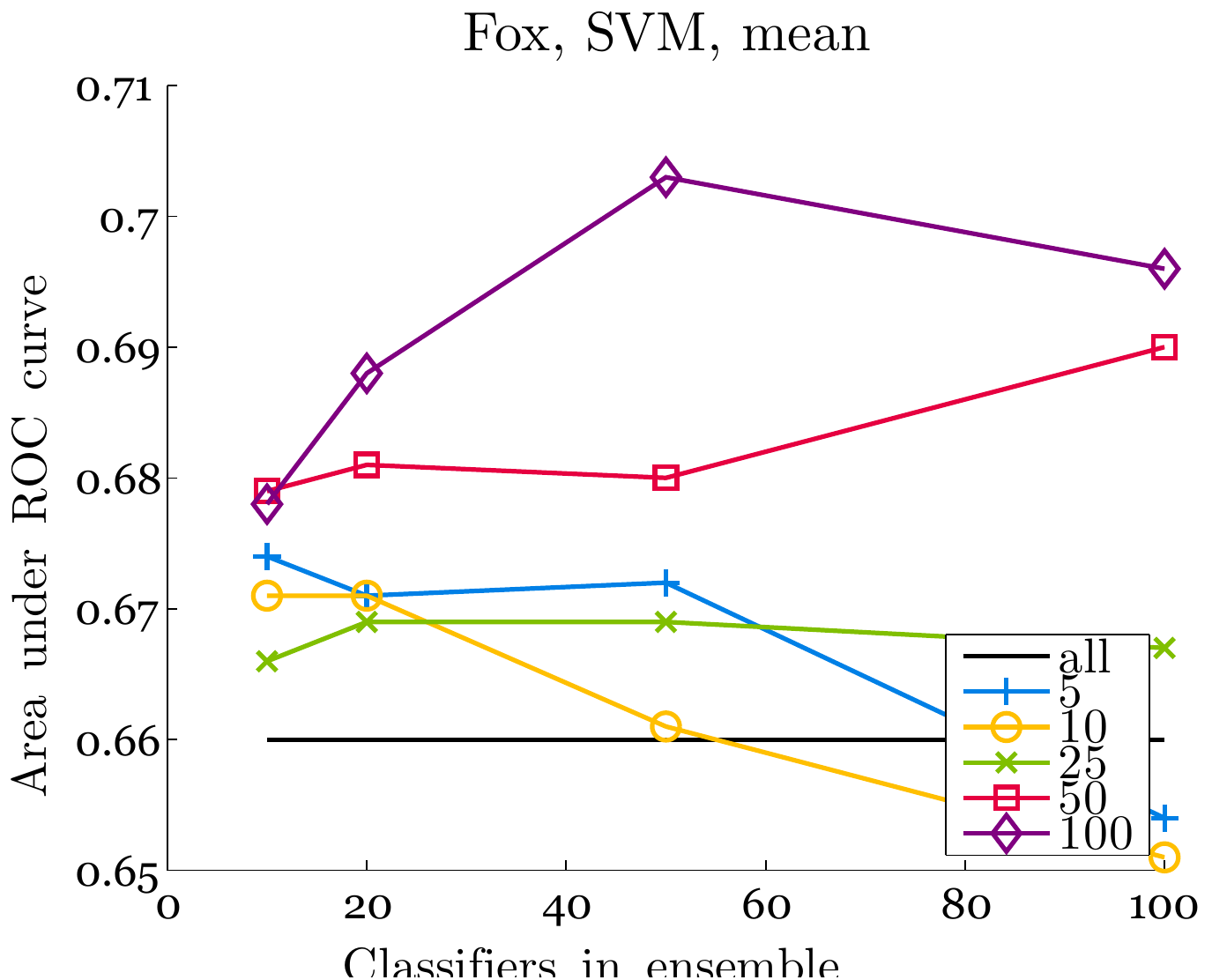}
   }
 \subfloat[]{
  \includegraphics[width=0.45\textwidth]{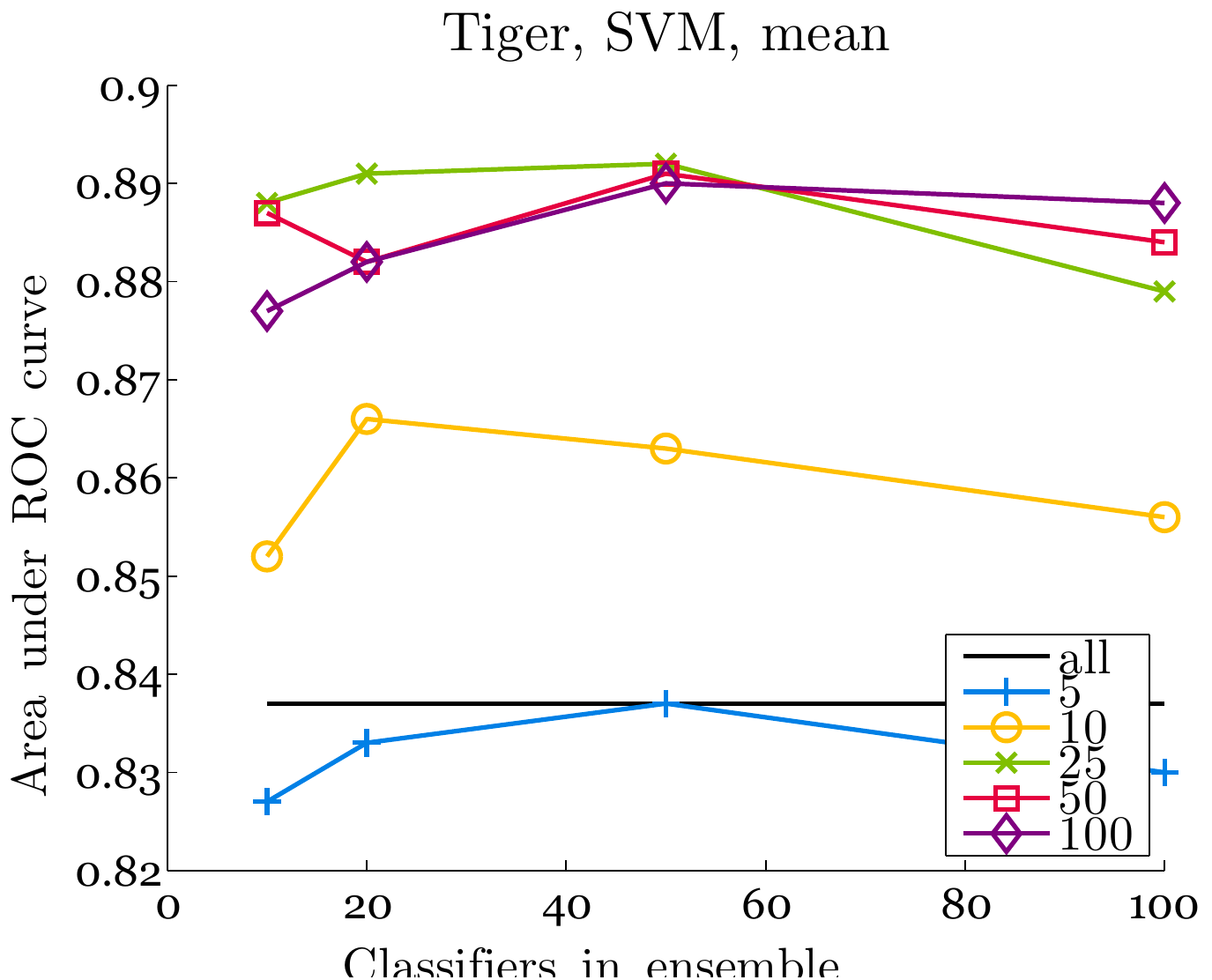}
   }
   
   \subfloat[]{
  \includegraphics[width=0.45\textwidth]{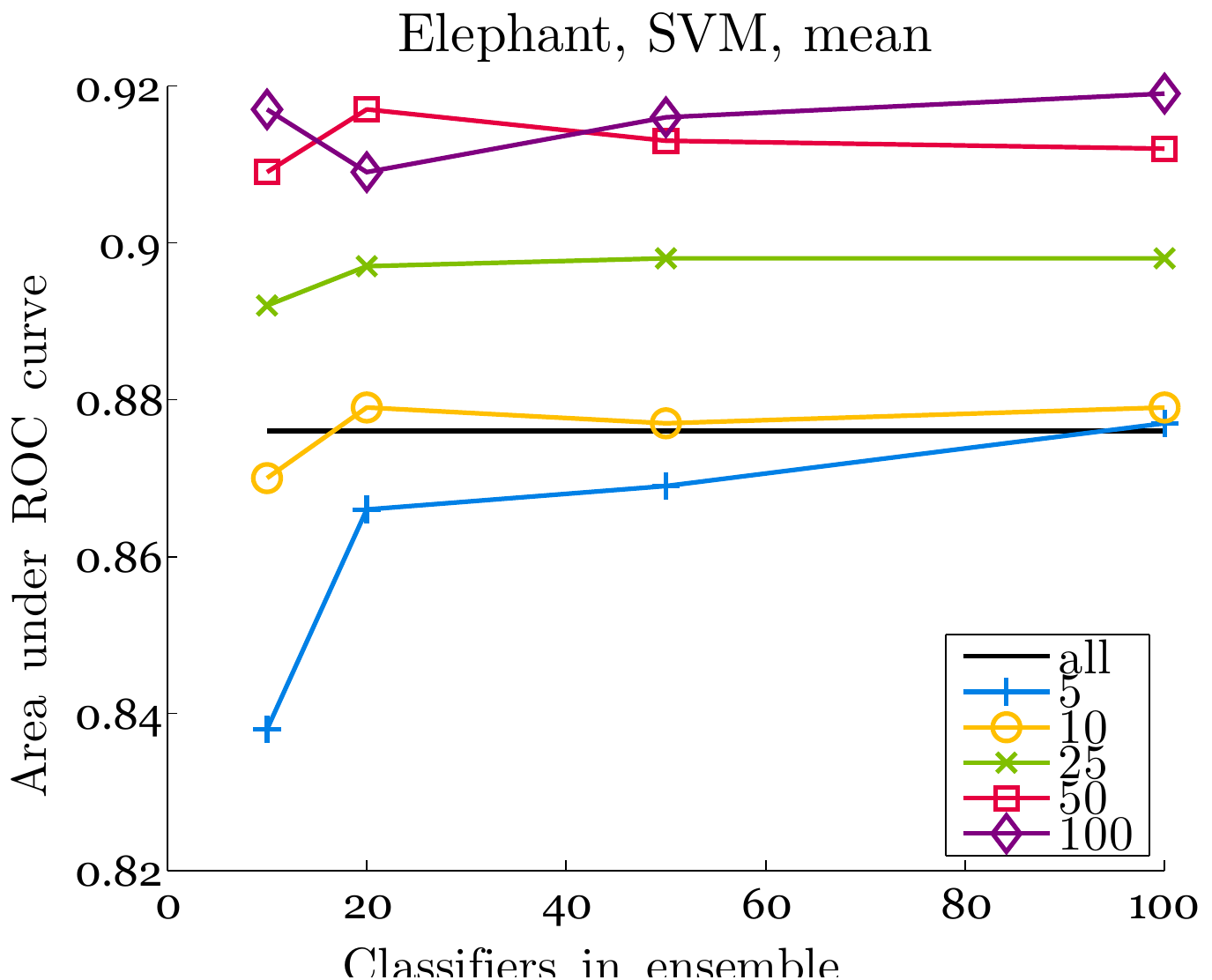}
   }
 \subfloat[]{
  \includegraphics[width=0.45\textwidth]{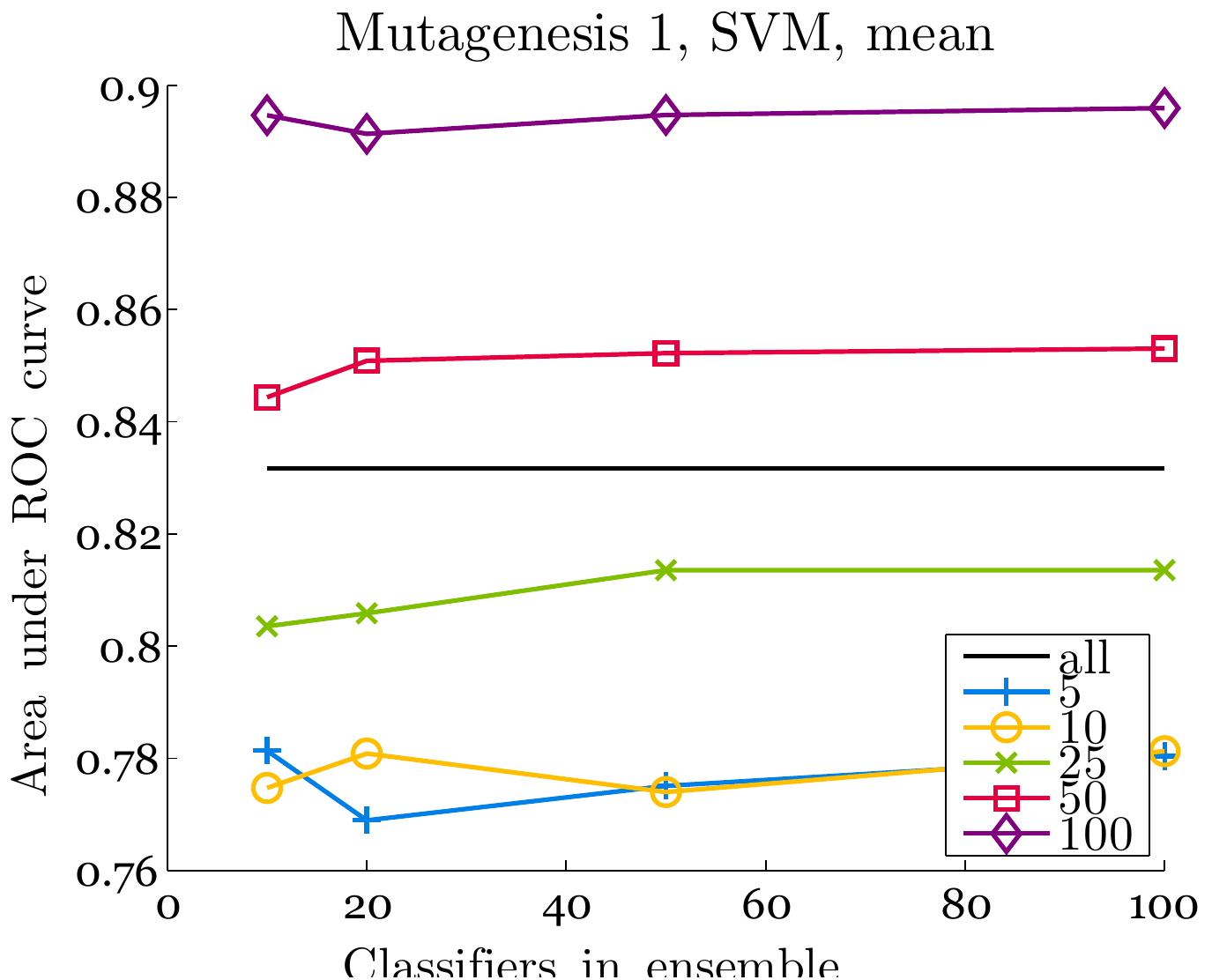}
   }

 \caption[]{AUC performances of the instance representation (black line) and the ensemble classifiers. Different lines per plot indicate different dimensionalities of subspace classifiers. }
 \label{fig:res_subspace1}
\end{figure*}

The results are shown in Fig.~\ref{fig:res_subspace1}. Here different plots show the results for different subspace dimensionalities (5, 10, 25, 50 or 100 features), and the x-axis indicates the number of classifiers used in the ensemble. The black line shows the baseline performance of $D^{inst}$. The performance metric is again the area under the ROC curve (AUC), and the results are averaged over 10-fold cross-validation. 


Ensembles created with higher-dimensional subspaces tend to perform better. An unsurprising result is that the fixed dimensionality values in this experiment are not suitable for all datasets. For example, in Musk none of the ensembles outperform the single classifier. Clearly, the subspace dimensionality should depend on the dimensionality (and possibly redundancy of the dissimilarities) of the original problem.

Another interesting observation in Fig.~\ref{fig:res_subspace1} is that it is often sufficient to build the ensemble from a few base classifiers, and adding further classifiers probably would not improve performance significantly. This is line with our earlier results for ensembles of one-class classifiers~\cite{cheplygina2011pruned}, even though both the data and the classifiers in question are very different. The recommendation in~\cite{kuncheva2010random} is that when there is no prior knowledge about how redundant the features are, $L$ should be chosen to be relatively small, while $s$ should be relatively large. 

Based on these observations, we settle on the following choices for $D^{RS}$: $L=100$ and $s=\frac{1}{5} \sum_i^N n_i$. We emphasize that these choices are good rules of thumb and do not have to be set to these exact values, which is supported by our results from the previous section: the performance is quite robust to changes in $L$ and $s$ provided $s$ is large enough. 

The performances of the proposed ensemble against those of the single classifier representations $D^{bag}$ and $D^{inst}$ are shown in Table~\ref{tab:disbased}. Contrary to our earlier results in~\cite{cheplygina2013combining}, $D^{RS}$ is now a clear competitor for the single classifiers, and has especially surprising performances on the Musk 1, Fox and Mutagenesis datasets. The advantages of $D^{RS}$ over the high-dimensional $D^{inst}$ are more visible, however, there are no significant differences with $D^{bag}$. This suggests that many of the dissimilarities are, in fact, informative, and averaging the dissimilarities over each bag preserves sufficient information. We believe this illustrates the strengths of the dissimilarity-based approach in general.

\begin{table}[ht]
\begin{center}
\begin{tabular}{l*{3}{c}}
& \multicolumn{3}{c}{Representation} \\
Dataset &      $D^{bag}$ &     $D^{inst}$ &       $D^{RS}$ \\ 
 \hline 
    Musk1 & {\bf 93.7 (3.5)}& {\bf 93.9 (3.6)}& {\bf 95.4 (2.4)}\\
    Musk2 & {\bf 95.7 (1.4)}& {\bf 89.7 (4.7)}& {\bf 93.2 (3.2)}\\
    Fox & {\bf 67.9 (2.5)}& {\bf 66.0 (3.9)}& {\bf 70.2 (1.8)}\\
    Tiger & {\bf 86.8 (5.2)}& 83.7 (3.9)& {\bf 87.8 (4.2)}\\
    Elephant & {\bf 90.9 (2.5)}& 87.6 (3.3)& {\bf 92.3 (2.7)}\\
    Mutagenesis & {\bf 84.3 (2.9)}& 83.2 (3.2)& {\bf 87.4 (3.5)}\\
    Brown Creeper & {\bf 94.7 (1.0)}& {\bf 93.9 (0.8)}& {\bf 94.2 (0.8)}\\
    Winter Wren & {\bf 99.5 (0.2)}& 98.4 (0.5)& {\bf 99.0 (0.3)}\\
    African & {\bf 92.5 (1.1)}& 91.6 (1.4)& {\bf 92.8 (1.2)}\\
    Beach & {\bf 88.3 (1.2)}& 85.6 (1.2)& {\bf 87.9 (1.2)}\\
    alt.atheism & {\bf 62.4 (8.3)}& {\bf 46.4 (5.9)}& {\bf 46.4 (5.2)}\\
   
\end{tabular}
\end{center}

\caption{AUC ($\times 100$) mean and standard error of 10-fold cross-validation of the single dissimilarity-based representations, and the proposed ensemble representation. Results in bold are best, or not significantly worse than best, per dataset}
\label{tab:disbased}
\end{table}


\subsection{Comparison with other MIL Classifiers}

We compare our method to other popular MIL classifiers, which cover a range of instance-based and bag-based methods, and are often being compared to in recent papers. 

EM-DD~\cite{zhang2001dd}, mi-SVM~\cite{andrews2002support} and MILBoost~\cite{viola2006multiple} are instance-based methods. EM-DD is an expectation-maximization algorithm which uses diverse density (DD), which, for a given point $t$ in the feature space space, measures the ratio between the number of positive bags which have instances near $t$, and the distance of the negative instances to $t$. The expectation step selects the most positive instance from each bag according to $t$, the maximization step then finds a new concept $t'$ by maximizing DD on the selected, most positive instances. mi-SVM is an extension of support vector machines which attempts to find hidden labels of the instances under constraints posed by the bag labels. Likewise, MILBoost is an extension of boosting, where the instances are reweighted in each of the boosting rounds. 

MILES~\cite{chen2006miles} and the Minimax kernel are bag-based methods which convert bags to a single-instance representation. MILES is similar to the 1-norm SVM applied to the $D^{inst}$, except that in MILES, a Gaussian kernel is used for defining similarities, and an appropriate $\sigma$ parameter is necessary. The Minimax kernel~\cite{gartner2002multi} is obtained by representing each bag by by the minimum and maximum feature values of its instances, this representation can then be used with a supervised classifier. All classifiers are implemented in PRTools~\cite{prtools} and the MIL toolbox~\cite{MIL2011} and default parameters are used unless stated otherwise. 

Next to the standard MIL classifiers, we use $D^{RS}$ with the guidelines from the previous section: the dimensionality of each subspace is 1/5th of the total dimensionality, and 100 classifiers are used in the ensemble. The base classifier is the linear SVM. The results are shown in Table~\ref{tab:comparison}. Some results could not be reported; in particular, for EM-DD when one cross-validation fold langs longer than five days, and MILBoost when features with the same value for all instances are present, as in alt.atheism. Needless to say, there is no method that performs the best at all times. Some of the default parameter choices may be unsuitable for a particular dataset, or the assumptions that the method is based on do not hold. However, overall the method we presented is always able to provide competitive performance.

\begin{table*}[ht]
\begin{center}

\begin{tabular}{l*{6}{c}}
& \multicolumn{6}{p{1.2cm}}{Classifier} \\
Dataset &     EM-DD & MI-SVM $r=10$ &  MILBoost & MILES $r=10$ & minimax+SVM &        $D^{RS}$+SVM  \\ 
 \hline

       Musk1 & 85.0 (5.1)& {\bf 91.5 (3.7)}& 74.8 (6.7)& {\bf 93.2 (2.9)}& {\bf 87.8 (5.0)}& {\bf 95.4 (2.4)}\\
       Musk2 & 88.1 (2.7)& {\bf 93.9 (2.8)}& 76.4 (3.5)& {\bf 97.1 (1.6)}& 91.3 (1.8)& {\bf 93.2 (3.2)}\\
       Fox & {\bf 67.6 (3.2)}& {\bf 68.7 (2.6)}& 61.3 (3.2)& {\bf 66.8 (3.5)}& 55.8 (2.9)& {\bf 70.2 (1.8)}\\
       Tiger & - & {\bf 87.2 (3.5)}& {\bf 87.0 (3.0)}& 84.6 (4.5)& 76.0 (4.1)& {\bf 87.8 (4.2)}\\
       Elephant & {\bf 88.5 (2.1)}& {\bf 90.7 (2.3)}& {\bf 88.8 (2.2)}& 88.4 (2.5)& {\bf 88.4 (2.1)}& {\bf 92.3 (2.7)}\\
       Mutagenesis & 67.4 (5.3)& 60.3 (4.5)& {\bf 88.1 (3.1)}& 72.1 (4.3)& 63.7 (4.4)& {\bf 87.4 (3.5)}\\
       Brown Creeper & 94.5 (0.9)& 92.8 (1.2)& {\bf 94.9 (0.9)}& {\bf 96.1 (0.6)}& 94.2 (0.9)& 94.2 (0.8)\\
       Winter Wren & {\bf 98.3 (0.5)}& {\bf 99.2 (0.4)}& {\bf 93.8 (5.6)}& {\bf 99.1 (0.5)}& 98.2 (0.3)& {\bf 99.0 (0.3)}\\
       African & {\bf 91.2 (1.8)}& 88.6 (1.7)& 89.4 (1.7)& 48.7 (2.3)& 87.6 (1.4)& {\bf 92.8 (1.2)}\\
       Beach & {\bf 84.6 (2.0)}& 78.2 (2.5)& {\bf 85.2 (2.9)}& 72.8 (5.0)& 83.0 (2.3)& {\bf 87.9 (1.2)}\\
       alt.atheism & 52.0 (8.0)& 38.8 (5.2)& -& 50.0 (5.5)& {\bf 80.0 (3.6)}& 46.4 (5.2)\\

       \end{tabular}
\end{center}

\caption{AUC $\times 100$, mean and standard error of 10-fold cross-validation of different MIL classifiers. Results in bold are best, or not significantly worse than best, per dataset. $r=10$ stands for radial basis kernel with width 10. }
\label{tab:comparison}
\end{table*}

\subsection{Instance weights}

\begin{figure}[ht!]
 \centering
 \subfloat[]{
  \includegraphics[width=0.90\columnwidth]{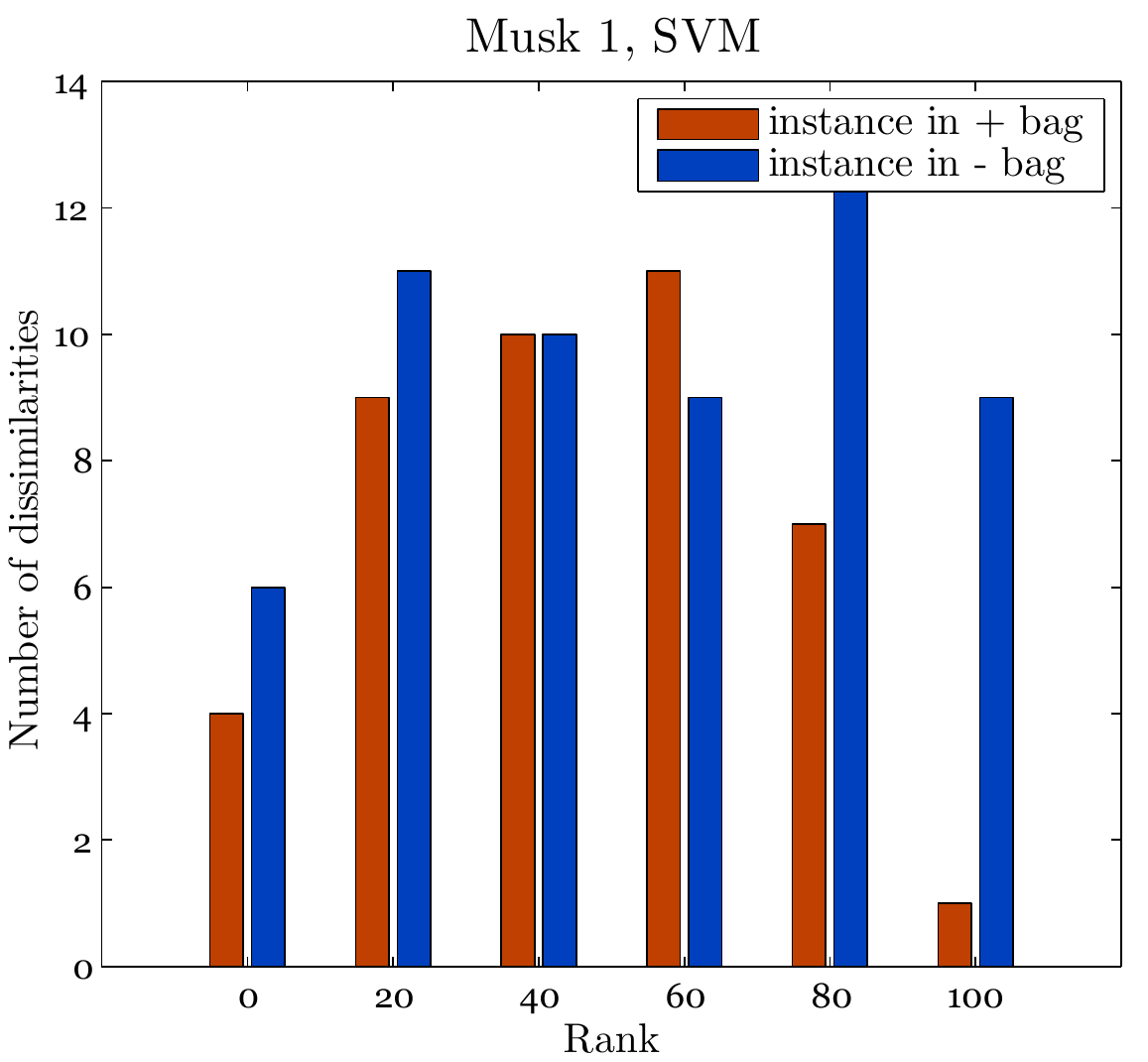}
   }
   
  \subfloat[]{
  \includegraphics[width=0.90\columnwidth]{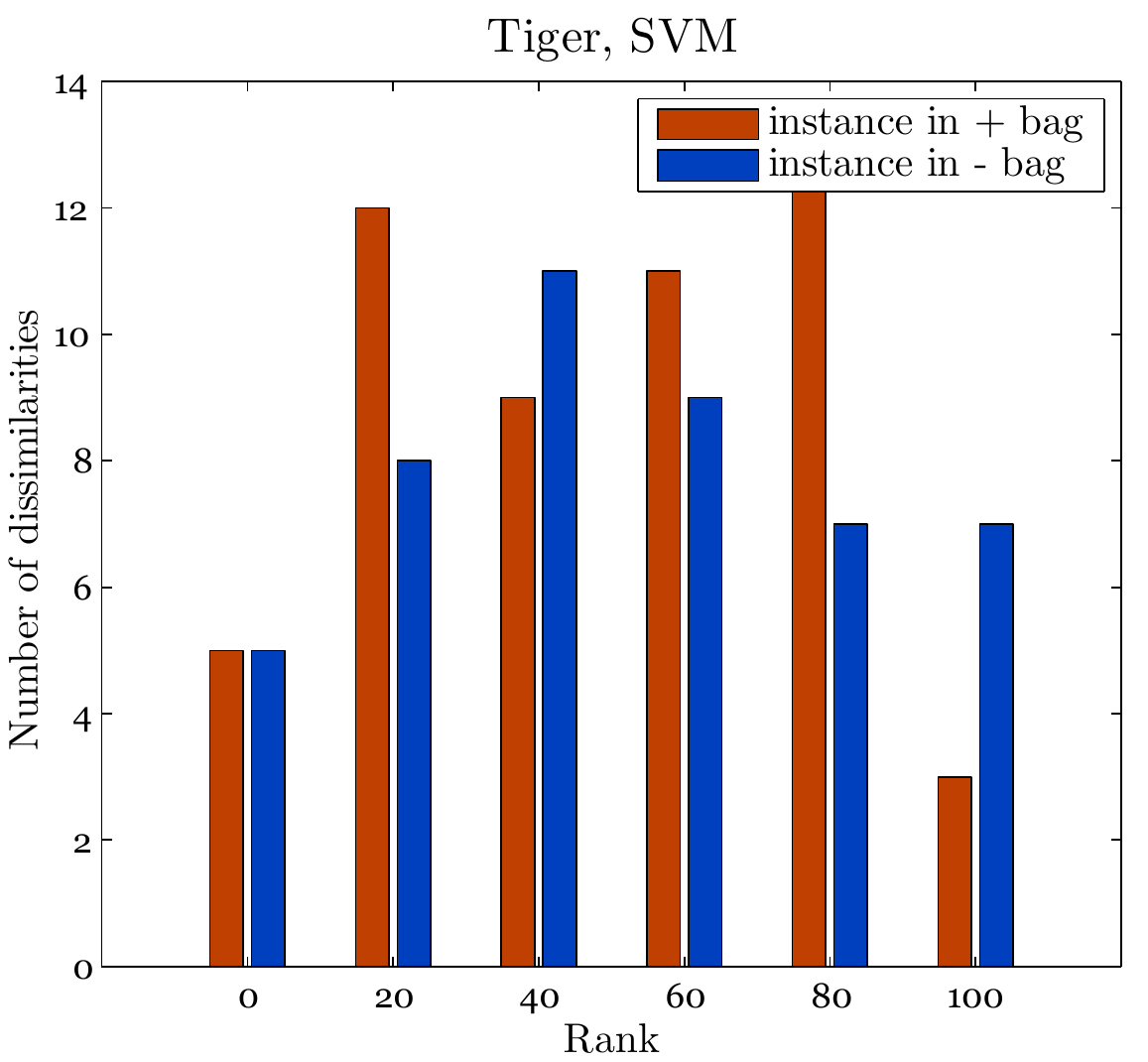}
   }

 \caption[]{Top 100 ranked dissimilarities, where the rank is determined by average weight of dissimilarities across $L=100$ subspace classifiers}
 \label{fig:ranks}
\end{figure}

An advantage of linear classifiers is the interpretability of the result --- from the weights $\mathbf{w}$ of the dissimilarities, we can derive which dissimilarities, and therefore which instances are more important (i.e., have a larger weight) to the classifier. This property can also be used in ensembles of linear classifiers, with a procedure described in~\cite{lai2006random}. For each dissimilarity, we calculate the average absolute value of its weight over all $L$ subspaces in which the dissimilarity was selected. We then sort the dissimilarities by this average weight, and view the position of each dissimilarity in this list as its rank. The distributions of the dissimilarities with ranks 1 to 100 are shown in Fig.\ref{fig:ranks} shows the distributions of top 100 dissimilarities. These most informative dissimilarities originate from both positive and negative bags, supporting the idea that not only concept, positive instances are important for these MIL problems.

\section{Discussion}\label{sec:conclusion}

We proposed a dissimilarity-based ensemble as a novel classification method for MIL problems. When bags are represented by their dissimilarities to instances from the training set, such instances can provide redundant information about the problem. A random subspace inspired ensemble, where classifiers are trained on different subspaces of the dissimilarity space, is a way of dealing with this redundancy. We show that our method achieves competitive performances with other MIL algorithms, and has intuitive parameters that do not need to be set by cross-validation to achieve these good results.

We investigated two choices for generating the subspaces: by using each training bag as a subspace, or by using a random selection of instances (with replacement) as a subspace. The random method achieved better results, especially when the dimensionality of the subspaces was increased. In fact, we found that the subspace dimensionality is the most important factor affecting the performance of the ensemble. On the other hand, the number of subspaces does not play a very important role and just a few classifiers are sufficient for good performance. These conclusions are in line with conclusions from other applications of the random subspace method, where the amount of redundancy of the features is unknown.

In general, the informativeness of a prototype is more related to the dimensionality of the subspace, then to the label of instances forming that subspaces. Negative bags, and unlabeled random sets of instances were often good prototypes, suggesting that most instances, and not only a few concept ones, are informative for these MIL problems. These results are more in line with the collective assumption
for MIL, where all instances are considered to contribute to the bag label, rather than with the standard assumption, where only a few positive instances are considered important.

Based on the encouraging results concerning the effectiveness of random subspaces as prototypes, we also considered randomly sampling the instance space (rather than randomly selecting existing instances) to generate artificial prototype bags that are not in the training set. Although the results with artificial prototypes were slightly worse than with real prototypes, this does seem to provide opportunities for using unlabeled, or artificial bags in a semi-supervised way.

We would like to conclude by emphasizing that a dissimilarity-based representation combined with a linear classifier (or an ensemble thereof) is a powerful way of classifying MIL bags. A question that still remains is the use of structured norms in such linear classifiers, which would enable selection of groups of dissimilarities, therefore revealing more about the relationships of the instances.


\bibliographystyle{IEEEtran}
\bibliography{refs}

\begin{thebibliography}{10}
\providecommand{\url}[1]{#1}
\csname url@samestyle\endcsname
\providecommand{\newblock}{\relax}
\providecommand{\bibinfo}[2]{#2}
\providecommand{\BIBentrySTDinterwordspacing}{\spaceskip=0pt\relax}
\providecommand{\BIBentryALTinterwordstretchfactor}{4}
\providecommand{\BIBentryALTinterwordspacing}{\spaceskip=\fontdimen2\font plus
\BIBentryALTinterwordstretchfactor\fontdimen3\font minus
  \fontdimen4\font\relax}
\providecommand{\BIBforeignlanguage}[2]{{%
\expandafter\ifx\csname l@#1\endcsname\relax
\typeout{** WARNING: IEEEtran.bst: No hyphenation pattern has been}%
\typeout{** loaded for the language `#1'. Using the pattern for}%
\typeout{** the default language instead.}%
\else
\language=\csname l@#1\endcsname
\fi
#2}}
\providecommand{\BIBdecl}{\relax}
\BIBdecl

\bibitem{dietterich1997solving}
T.~G. Dietterich, R.~H. Lathrop, and T.~Lozano-P{\'e}rez, ``Solving the
  multiple instance problem with axis-parallel rectangles,'' \emph{Artificial
  Intelligence}, vol.~89, no. 1-2, pp. 31--71, 1997.

\bibitem{fu2012implementation}
G.~Fu, X.~Nan, H.~Liu, R.~Patel, P.~Daga, Y.~Chen, D.~Wilkins, and R.~Doerksen,
  ``Implementation of multiple-instance learning in drug activity prediction,''
  \emph{BMC Bioinformatics}, vol.~13, no. Suppl 15, p.~S3, 2012.

\bibitem{andrews2002multiple}
S.~Andrews, T.~Hofmann, and I.~Tsochantaridis, ``Multiple instance learning
  with generalized support vector machines,'' in \emph{National Conference on
  Artificial Intelligence}, 2002, pp. 943--944.

\bibitem{chen2006miles}
Y.~Chen, J.~Bi, and J.~Z. Wang, ``Miles: Multiple-instance learning via
  embedded instance selection,'' \emph{IEEE Transactions on Pattern Analysis
  and Machine Intelligence}, vol.~28, no.~12, pp. 1931--1947, 2006.

\bibitem{zhou2009multi}
Z.~H. Zhou, Y.~Y. Sun, and Y.~F. Li, ``Multi-instance learning by treating
  instances as non-iid samples,'' in \emph{International Conference on Machine
  Learning}.\hskip 1em plus 0.5em minus 0.4em\relax ACM, 2009, pp. 1249--1256.

\bibitem{fung2007multiple}
G.~Fung, M.~Dundar, B.~Krishnapuram, and R.~B. Rao, ``Multiple instance
  learning for computer aided diagnosis,'' in \emph{Advances in Neural
  Information Processing Systems}, vol.~19, 2007, p. 425.

\bibitem{foulds2010review}
J.~Foulds and E.~Frank, ``A review of multi-instance learning assumptions,''
  \emph{Knowledge Engineering Review}, vol.~25, no.~1, p.~1, 2010.

\bibitem{wang2000solving}
J.~Wang, ``Solving the multiple-instance problem: A lazy learning approach,''
  in \emph{International Conference on Machine Learning}, 2000.

\bibitem{gartner2002multi}
T.~G{\"a}rtner, P.~A. Flach, A.~Kowalczyk, and A.~J. Smola, ``Multi-instance
  kernels,'' in \emph{International Conference on Machine Learning}, 2002, pp.
  179--186.

\bibitem{tax2011bag}
D.~M.~J. Tax, M.~Loog, R.~P.~W. Duin, V.~Cheplygina, and W.-J. Lee, ``Bag
  dissimilarities for multiple instance learning,'' in \emph{Similarity-Based
  Pattern Recognition}.\hskip 1em plus 0.5em minus 0.4em\relax Springer, 2011,
  pp. 222--234.

\bibitem{pekalska2005dissimilarity}
E.~P{\k{e}}kalska and R.~P.~W. Duin, \emph{The dissimilarity representation for
  pattern recognition: foundations and applications}.\hskip 1em plus 0.5em
  minus 0.4em\relax World Scientific Pub Co Inc, 2005, vol.~64.

\bibitem{cheplygina2012class}
V.~Cheplygina, D.~M.~J. Tax, and M.~Loog, ``Class-dependent dissimilarity
  measures for multiple instance learning,'' in \emph{Structural, Syntactic,
  and Statistical Pattern Recognition}.\hskip 1em plus 0.5em minus 0.4em\relax
  Springer, 2012, pp. 602--610.

\bibitem{pekalska2001combining}
E.~P{\k{e}}kalska and R.~P.~W. Duin, ``On combining dissimilarity
  representations,'' in \emph{Multiple Classifier Systems}.\hskip 1em plus
  0.5em minus 0.4em\relax Springer, 2001, pp. 359--368.

\bibitem{kittler1998combining}
J.~Kittler, ``{Combining classifiers: A theoretical framework},'' \emph{Pattern
  Analysis \& Applications}, vol.~1, no.~1, pp. 18--27, 1998.

\bibitem{duin2000experiments}
R.~P.~W. Duin and D.~M.~J. Tax, ``{Experiments with classifier combining
  rules},'' in \emph{Multiple Classifier Systems}.\hskip 1em plus 0.5em minus
  0.4em\relax Springer, 2000, pp. 16--29.

\bibitem{cheplygina2013combining}
V.~Cheplygina, D.~M.~J. Tax, and M.~Loog, ``Combining instance information to
  classify bags,'' in \emph{Multiple Classifier Systems}.\hskip 1em plus 0.5em
  minus 0.4em\relax Springer, 2013, pp. 13--24.

\bibitem{zhou2006multi}
Z.-H. Zhou and M.-L. Zhang, ``Multi-instance multi-label learning with
  application to scene classification,'' in \emph{Advances in Neural
  Information Processing Systems}, 2006, pp. 1609--1616.

\bibitem{cheplygina2012does}
V.~Cheplygina, D.~M.~J. Tax, and M.~Loog, ``Does one rotten apple spoil the
  whole barrel?'' in \emph{International Conference on Pattern
  Recognition}.\hskip 1em plus 0.5em minus 0.4em\relax IEEE, 2012, pp.
  1156--1159.

\bibitem{lai2006random}
C.~Lai, M.~J.~T. Reinders, and L.~Wessels, ``Random subspace method for
  multivariate feature selection,'' \emph{Pattern Recognition Letters},
  vol.~27, no.~10, pp. 1067--1076, 2006.

\bibitem{zhu20041}
J.~Zhu, S.~Rosset, T.~Hastie, and R.~Tibshirani, ``1-norm support vector
  machines,'' in \emph{Advances in Neural Information Processing Systems},
  vol.~16, no.~1.\hskip 1em plus 0.5em minus 0.4em\relax The MIT Press, 2004,
  pp. 49--56.

\bibitem{bhattacharyya2003simultaneous}
C.~Bhattacharyya, L.~Grate, A.~Rizki, D.~Radisky, F.~Molina, M.~I. Jordan,
  M.~J. Bissell, and I.~S. Mian, ``Simultaneous classification and relevant
  feature identification in high-dimensional spaces: application to molecular
  profiling data,'' \emph{Signal Processing}, vol.~83, no.~4, pp. 729--743,
  2003.

\bibitem{ho1998random}
T.~Ho, ``{The random subspace method for constructing decision forests},''
  \emph{IEEE Transactions on Pattern Analysis and Machine Intelligence},
  vol.~20, no.~8, pp. 832--844, 1998.

\bibitem{kuncheva2003measures}
L.~I. Kuncheva and C.~J. Whitaker, ``Measures of diversity in classifier
  ensembles and their relationship with the ensemble accuracy,'' \emph{Machine
  Learning}, vol.~51, no.~2, pp. 181--207, 2003.

\bibitem{brown2005diversity}
G.~Brown, J.~Wyatt, R.~Harris, and X.~Yao, ``Diversity creation methods: a
  survey and categorisation,'' \emph{Information Fusion}, vol.~6, no.~1, pp.
  5--20, 2005.

\bibitem{skurichina2001bagging}
M.~Skurichina and R.~P.~W. Duin, ``Bagging and the random subspace method for
  redundant feature spaces,'' in \emph{Multiple Classifier Systems}.\hskip 1em
  plus 0.5em minus 0.4em\relax Springer, 2001, pp. 1--10.

\bibitem{kuncheva2010random}
L.~I. Kuncheva, J.~J. Rodr{\'\i}guez, C.~O. Plumpton, D.~E. Linden, and S.~J.
  Johnston, ``Random subspace ensembles for fmri classification,'' \emph{IEEE
  Transactions on Medical Imaging}, vol.~29, no.~2, pp. 531--542, 2010.

\bibitem{bertoni2005bio}
A.~Bertoni, R.~Folgieri, and G.~Valentini, ``Bio-molecular cancer prediction
  with random subspace ensembles of support vector machines,''
  \emph{Neurocomputing}, vol.~63, pp. 535--539, 2005.

\bibitem{ham2005investigation}
J.~Ham, Y.~Chen, M.~M. Crawford, and J.~Ghosh, ``Investigation of the random
  forest framework for classification of hyperspectral data,'' \emph{IEEE
  Transactions on Geoscience and Remote Sensing}, vol.~43, no.~3, pp. 492--501,
  2005.

\bibitem{maron1998framework}
O.~Maron and T.~Lozano-P{\'e}rez, ``A framework for multiple-instance
  learning,'' in \emph{Advances in Neural Information Processing Systems},
  1998, pp. 570--576.

\bibitem{srinivasan1995comparing}
A.~Srinivasan, S.~Muggleton, and R.~King, ``Comparing the use of background
  knowledge by inductive logic programming systems,'' in \emph{International
  Workshop on Inductive Logic Programming}, 1995, pp. 199--230.

\bibitem{briggs2012acoustic}
F.~Briggs, B.~Lakshminarayanan, L.~Neal, X.~Fern, R.~Raich, S.~Hadley,
  A.~Hadley, and M.~Betts, ``Acoustic classification of multiple simultaneous
  bird species: A multi-instance multi-label approach,'' \emph{Journal of the
  Acoustical Society of America}, vol. 131, p. 4640, 2012.

\bibitem{chapelle2007training}
O.~Chapelle, ``Training a support vector machine in the primal,'' \emph{Neural
  Computation}, vol.~19, no.~5, pp. 1155--1178, 2007.

\bibitem{tax2000combining}
D.~M.~J. Tax, M.~Van~Breukelen, R.~P.~W. Duin, and J.~Kittler, ``Combining
  multiple classifiers by averaging or by multiplying?'' \emph{Pattern
  Recognition}, vol.~33, no.~9, pp. 1475--1485, 2000.

\bibitem{cheplygina2011pruned}
V.~Cheplygina and D.~M.~J. Tax, ``Pruned random subspace method for one-class
  classifiers,'' in \emph{Multiple Classifier Systems}.\hskip 1em plus 0.5em
  minus 0.4em\relax Springer, 2011, pp. 96--105.

\bibitem{bradley1997use}
A.~P. Bradley, ``{The use of the area under the {ROC} curve in the evaluation
  of machine learning algorithms},'' \emph{Pattern Recognition}, vol.~30,
  no.~7, pp. 1145--1159, 1997.

\bibitem{huang2005using}
J.~Huang and C.~X. Ling, ``Using {AUC} and accuracy in evaluating learning
  algorithms,'' \emph{IEEE Transactions on Knowledge and Data Engineering},
  vol.~17, no.~3, pp. 299--310, 2005.

\bibitem{tax2008learning}
D.~M.~J. Tax and R.~P.~W. Duin, ``Learning curves for the analysis of multiple
  instance classifiers,'' in \emph{Structural, Syntactic, and Statistical
  Pattern Recognition}.\hskip 1em plus 0.5em minus 0.4em\relax Springer, 2008,
  pp. 724--733.

\bibitem{dinh2012study}
C.~V. Dinh, R.~P.~W. Duin, and M.~Loog, ``A study on semi-supervised
  dissimilarity representation,'' in \emph{International Conference on Pattern
  Recognition}.\hskip 1em plus 0.5em minus 0.4em\relax IEEE, 2012, pp.
  2861--2864.

\bibitem{pkekalska2002discussion}
E.~P{\k{e}}kalska, R.~P.~W. Duin, and M.~Skurichina, ``A discussion on the
  classifier projection space for classifier combining,'' in \emph{Multiple
  Classifier Systems}.\hskip 1em plus 0.5em minus 0.4em\relax Springer, 2002,
  pp. 137--148.

\bibitem{brown2010good}
G.~Brown and L.~I. Kuncheva, ``"{G}ood" and "bad" diversity in majority vote
  ensembles,'' in \emph{Multiple Classifier Systems}.\hskip 1em plus 0.5em
  minus 0.4em\relax Springer, 2010, pp. 124--133.

\bibitem{zhang2001dd}
Q.~Zhang, S.~Goldman \emph{et~al.}, ``Em-dd: An improved multiple-instance
  learning technique,'' in \emph{Advances in Neural Information Processing
  Systems}, vol.~14.\hskip 1em plus 0.5em minus 0.4em\relax Cambridge, MA: MIT
  Press, 2001, pp. 1073--1080.

\bibitem{andrews2002support}
S.~Andrews, I.~Tsochantaridis, and T.~Hofmann, ``Support vector machines for
  multiple-instance learning,'' in \emph{Advances in Neural Information
  Processing Systems}, vol.~15, 2002, pp. 561--568.

\bibitem{viola2006multiple}
P.~Viola, J.~Platt, and C.~Zhang, ``Multiple instance boosting for object
  detection,'' in \emph{Advances in Neural Information Processing Systems},
  vol.~18.\hskip 1em plus 0.5em minus 0.4em\relax Citeseer, 2006, p. 1417.

\bibitem{prtools}
R.~P.~W. Duin, P.~Juszczak, P.~Paclik, E.~Pekalska, D.~De~Ridder, D.~M.~J. Tax,
  and S.~Verzakov, ``Prtools, a matlab toolbox for pattern recognition,''
  online, http://www.prtools.org, 2013.

\bibitem{MIL2011}
\BIBentryALTinterwordspacing
D.~M.~J. Tax, ``{MIL}, a {M}atlab toolbox for multiple instance learning,'' May
  2011, version 0.7.9. [Online]. Available:
  \url{http://prlab.tudelft.nl/david-tax/mil.html}
\BIBentrySTDinterwordspacing

\end{thebibliography}

\begin{IEEEbiography}[{\includegraphics[width=1in,
height=1.25in,clip,keepaspectratio]{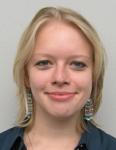}}]{Veronika Cheplygina} received her M.Sc. degree in Media and Knowledge Engineering from the Delft University of Technology, the Netherlands in 2010. Her thesis project ``Random Subspace Method for One-class Classifiers'' about detecting outliers during automatic parcel sorting was performed in collaboration with Prime Vision (\url{http://www.primevision.com/}). She is currently working towards her Ph.D. at the Pattern Recognition Laboratory at the Delft University of Technology. Her research interests include multiple instance learning, dissimilarity representation, learning in (non)-metric spaces, and structured data.
\end{IEEEbiography}
\begin{IEEEbiography}[{\includegraphics[width=1in,
height=1.25in,clip,keepaspectratio]{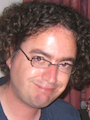}}]{David M.J. Tax} studied Physics at the University of Nijmegen, the Netherlands in 1996, and received his Masters degree with the thesis ``Learning of Structure by Many-take-all Neural Networks''. After that he received his Ph.D. with the thesis ``One-class Classification'' from the Delft University of Technology, the Netherlands, under the supervision of Dr. Robert P.W. Duin. After working for two years as a MarieCurie Fellow in the Intelligent Data Analysis group in Berlin, he is currently an assistant professor in the Pattern Recognition Laboratory at the Delft University of Technology. His main research interest is in the learning and development of detection algorithms and (one-class) classifiers that optimize alternative performance criteria like ordering criteria using the Area under the ROC curve or a Precision-Recall graph.
Furthermore, the problems concerning the representation of data, multiple instance learning, simple and elegant classifiers and the fair evaluation of methods have focus.
\end{IEEEbiography}
\begin{IEEEbiography}[{\includegraphics[width=1in,
height=1.25in,clip,keepaspectratio]{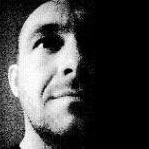}}]{Marco Loog} received an M.Sc. degree in mathematics from Utrecht University and in 2004 a Ph.D. degree from the Image Sciences Institute for the development and improvement of contextual statistical pattern recognition methods and their use in the processing and analysis of images. After this joyful event, he moved to Copenhagen where he acted as assistant and, eventually, associate professor next to which he worked as a research scientist at Nordic Bioscience. In 2008, after several splendid years in Denmark, Marco moved to Delft University of Technology where he now works as an assistant professor in the Pattern Recognition Laboratory. He currently is chair of Technical Committee 1 of the IAPR, holds a bunch of associate editorships, is honorary full professor in pattern recognition at the University of Copenhagen, and is also affiliated to Eindhoven University of Technology. Marco's research interests include multiscale image analysis, semi-supervised and multiple instance learning, saliency, computational perception, the dissimilarity approach, and black math.
\end{IEEEbiography}
\vfill
%








\end{document}